%% file: main.tex
\documentclass[10pt]{article}
\usepackage[preprint]{tmlr}

\usepackage[T1]{fontenc}
\usepackage{amsmath,amssymb}
\usepackage{booktabs}
\usepackage{wrapfig}
\usepackage{etoc}
\usepackage{graphicx}
\usepackage{xcolor}
\usepackage{microtype}
\usepackage{hyperref}
\definecolor{iclrblue}{rgb}{0.18, 0.46, 0.71}
\hypersetup{
    colorlinks=true,
    linkcolor=iclrblue,
    citecolor=iclrblue,
    urlcolor=iclrblue
}
\usepackage{url}
\usepackage[capitalize,noabbrev]{cleveref}
\usepackage{pifont}
\usepackage{enumitem}
\setlist{itemsep=1pt,topsep=3pt,leftmargin=*}
\usepackage[acronym]{glossaries}
\glsdisablehyper
\setacronymstyle{long-short}
\newacronym{bo}{BO}{Bayesian Optimization}
\newacronym{gp}{GP}{Gaussian Process}
\newacronym{hpo}{HPO}{Hyperparameter Optimization}
\newacronym{rr}{RR}{Relax-and-Round}
\newacronym{cnn}{CNN}{Convolutional Neural Network}
\newacronym{pr}{PR}{Probabilistic Reparameterization}
\newacronym{ei}{EI}{Expected Improvement}
\newacronym{logei}{LogEI}{log Expected Improvement}
\newacronym{qlognei}{qLogNEI}{batch log Noisy Expected Improvement}
\newacronym{ucb}{UCB}{Upper Confidence Bound}
\newacronym{tpe}{TPE}{Tree-structured Parzen Estimator}
\newacronym{smac}{SMAC}{Sequential Model-based Algorithm Configuration}
\newacronym{cocabo}{CoCaBO}{Continuous and Categorical BO}
\newacronym{ard}{ARD}{Automatic Relevance Determination}
\newacronym{cd}{CD}{Critical Difference}
\newacronym{rope}{ROPE}{Region Of Practical Equivalence}
\newacronym{bca}{BCa}{bias-corrected and accelerated}
\newacronym{mlp}{MLP}{Multi-Layer Perceptron}
\newacronym{gpu}{GPU}{Graphics Processing Unit}

\usepackage{xspace}
\newcommand{\mRRAx}{\mbox{RR-Ax}\xspace}
\newcommand{\mRRext}{\mbox{RR-ext}\xspace}
\newcommand{\mRRcache}{\mbox{RR+cache}\xspace}
\newcommand{\mRRdedup}{\mbox{RR+dedup}\xspace}
\newcommand{\mKRound}{\mbox{KernelRound}\xspace}
\newcommand{\mPR}{PR\xspace}
\newcommand{\mDisc}{Discrete\xspace}
\newcommand{\mAxDef}{\mbox{Ax-default}\xspace}
\newcommand{\mRRnb}{\mbox{RR+neighbour}\xspace}
\newcommand{\one}{\ding{182}}
\newcommand{\two}{\ding{183}}
\newcommand{\three}{\ding{184}}

\title{Bayesian Optimization for Hyperparameters Tuning\\in Neural Networks}
\author{
      \name Gabriele Onorato  \\
      \addr Sapienza University of Rome
      \AND
      \name Giampaolo Liuzzi \\
      \addr Sapienza University of Rome
      }

\def\month{11}
\def\year{2026}
\def\openreview{\url{https://openreview.net/forum?id=XXXX}}

\begin{document}

\maketitle

\begin{abstract}
While the performance of a neural network depends on integer hyperparameters such as layer widths,
depth and batch size, finding a good configuration is costly, since every evaluation requires a full training run. One commonly used technique to address this cost is \gls{bo}, which, with a \gls{gp} surrogate,
is suited to such expensive, black-box, non-differentiable objectives. Often, these integer hyperparameters are handled by relaxing them to real
numbers and rounding the acquisition optimum. However, this \gls{rr} strategy can re-propose configurations
it has already evaluated, wasting training runs. We study when \gls{rr} actually hurts and what its alternatives cost, comparing
eight integer-handling variants in Ax/BoTorch, seven sharing one surrogate, with seven \gls{hpo}
baselines on 36 synthetic tasks, 20 tabular benchmarks of neural network \gls{hpo} and a CIFAR-10
\gls{cnn}, for about 31{,}000 paired runs. Our experiments indicate that \one\ \gls{rr} barely changes
final regret but delays the search by about four evaluations, because the relaxed acquisition often
peaks inside the cell of an evaluated point, and neither Ax's native deduplication nor, on our
two-dimensional tasks, replacing such proposals with the best unobserved neighbouring grid point
removes the delay; \two\ \gls{rr} wastes up to 49\% of its trials on coarse grids and 63\% under
an exploitative acquisition, mostly because it does not exclude observed points, with rounding adding up to
about 7 percentage points; and \three\ the \gls{gp} lengthscale modulates the waste, but neither
fixing nor bounding it removes it. We also identify duplicates at boundary optima of mixed spaces,
which only the exclusion of observed points avoids. Finally, on our CIFAR-10 search spaces, every method ends within 0.35 percentage points of test accuracy, about
the training-seed noise. Hence, optimizing the acquisition over the grid avoids the delay, and
excluding observed points removes the waste at no cost in regret.
\end{abstract}
\glsresetall

\section{Introduction}
\label{sec:intro}

The performance of a neural network on a given task depends on hyperparameters that are fixed before training,
such as the width and number of layers, the batch size and the learning rate
\citep{bergstra2011algorithms, bischl2023hyperparameter}. Since evaluating a single
configuration requires training the network, \gls{hpo} aims to find a
good configuration within a small budget of training runs. In this setting, \gls{bo} with
\gls{gp} surrogates is a standard tool \citep{snoek2012practical,
shahriari2016taking, turner2021bayesian}: it models the validation accuracy as a function of the
hyperparameters through a \gls{gp}, whose covariance function, or kernel (\cref{sec:bo}), encodes how similar the accuracies of two configurations are expected to be as a function of their distance, and selects each new configuration by maximizing an acquisition function.

However, many of these hyperparameters are integers, e.g., the number of units or channels of a
layer, the number of layers and the batch size. Since standard kernels and gradient-based acquisition optimizers operate on
$\mathbb{R}^d$, a common practical treatment \emph{relaxes} each integer to a real
interval, then fits the \gls{gp} and optimizes the acquisition function in the continuous box, and finally
\emph{rounds} the proposed point before evaluating it \citep{garridomerchan2020dealing,
luong2019bayesian}. We refer to this strategy as
\gls{rr}. Several alternatives to \gls{rr} exist, from rounding inside the kernel
\citep{garridomerchan2020dealing} to \gls{pr} of the acquisition
\citep{daulton2022bayesian} and combinatorial or mixed-variable \gls{bo} \citep{baptista2018bayesian,
oh2019combinatorial, ru2020bayesian, wan2021think}. To address this issue, libraries have changed as well: the
current Ax/BoTorch software (Ax 1.3.1, \citealp{olson2025ax, balandat2020botorch}) keeps integers discrete and excludes
observed points by default, while Ax versions before 1.0 relaxed them. However, \gls{rr} remains common in practice. Custom \gls{bo} loops built on continuous acquisition optimizers relax integers, and so do batch proposals and trust-region methods such as TuRBO \citep{eriksson2019scalable}. HEBO \citep{cowenrivers2022hebo} also optimizes its acquisition over the relaxed range and rounds. As we show, excluding observed points is needed even when the acquisition is optimized on the grid.

\begin{figure}[t]
\centering
\includegraphics[width=\textwidth]{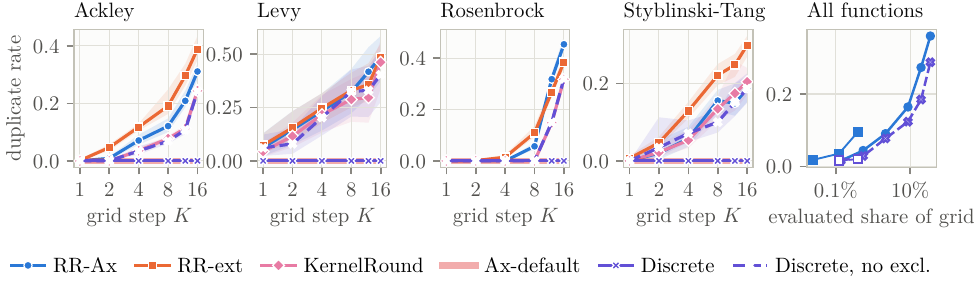}\vspace{-10pt}
\caption{\textbf{Duplicates Grow with the Grid Step.} \textbf{\emph{Left:}} duplicate rate of the \acrshort{bo} phase against the grid step $K$ of a fixed physical range
($d=2$, mean and 95\% \acrshort{bca} band over 20 seeds). The relaxed variants \mRRAx, \mRRext and \mKRound re-propose
up to half of their trials as the grid gets coarser, while \mDisc and \mAxDef never re-propose a point.
\mDisc without the exclusion of observed points (dashed) also re-proposes, on average up to about 7 percentage points below \mRRAx: most
duplicates come from not excluding observed points, and rounding adds the rest. \textbf{\emph{Right:}} the same rates against the share of the grid evaluated within the budget (mean over functions; circles for $d=2$, squares for $d=4$, per function in \cref{fig:dupk_full}).\vspace{-10pt}}
\label{fig:dupk}
\end{figure}

\noindent\textbf{Key Issue.}
\gls{rr} has two distinct costs. The first is \emph{waste}: under \gls{rr}, the \gls{gp} treats points
between grid values as informative, while the objective cannot distinguish them, so the acquisition
optimum can land inside the cell of an evaluated configuration and be rounded back onto it.
However, rounding is not the only source of duplicates. On a small, coarse grid, the acquisition
maximum is often an evaluated point itself, so an optimizer that does not exclude observed points
re-proposes it even without any relaxation. As shown in \Cref{fig:dupk}, with a fixed physical
range, the duplicate rate of \gls{rr} grows from almost zero to 31\% on Ackley and 49\% on Levy as
the grid step increases, even with a modern acquisition function and inferred noise, while discrete
acquisition optimization without exclusion stays on average up to about 7 percentage points (pp) lower, and a
single \gls{rr} run proposes the same configuration up to 34 times. The second cost is \emph{delay}:
\gls{rr} needs more evaluations to reach a good configuration, even when Ax rejects duplicates, since
the relaxed acquisition often peaks inside the cell of an evaluated point, and Ax then re-evaluates
it or falls back on a worse point.

There are several technical challenges to disentangle this effect and establish whether \gls{rr} actually hurts. First, the integer treatment is
usually entangled with other choices, such as the surrogate, the acquisition function and the
kind of implementation (e.g., the library version), so that comparisons across papers or tools cannot attribute differences to it.
Second, the waste depends on properties of the task (grid step, dimension, position of the
optimum) and of the optimizer (acquisition function, noise model, initial design, batch size),
which have not been varied systematically. Finally, the accuracy differences between
hyperparameter configurations of neural networks are often of the same order as training-seed
noise. Consequently, single-seed results selected on the test set cannot separate them.

\noindent\textbf{Proposed Approach.}
To address the key issue described above, we conduct a controlled empirical study that isolates
the integer treatment. To do so, we compare seven integer-handling variants that share one
\gls{gp} surrogate and one acquisition function and differ only in how integers enter the
\gls{bo} loop: \gls{rr} as in Ax versions before 1.0, \gls{rr} outside the model, caching and
deduplication, rounding inside the kernel, \gls{pr}, and exhaustive discrete
acquisition optimization. The current Ax default is our reference, and seven \gls{hpo} baselines complete the comparison. Crucially, every method receives the same Sobol initial design for each seed, so that every comparison is paired. We also vary one factor at a time to isolate what drives the waste. Finally, we tune a CIFAR-10 \gls{cnn} with a
validation split, test-only final evaluation and five runs per method.
\smallskip

\noindent\textbf{Summary of Contributions:}
\begin{enumerate}[topsep=0pt]
\item \textbf{Controlled Comparison:} We benchmark eight integer-handling variants and seven
  baselines on 36 synthetic tasks, 20 tabular \gls{hpo} tasks, three JAHS-Bench-201 tasks and two
  CIFAR-10 search spaces, for about 31{,}000 paired runs (\cref{sec:methods,sec:setup}).
\item \textbf{Final Quality:} We show that final regret is largely insensitive to the integer
  treatment, since \gls{rr} and the Ax default are practically equivalent on the synthetic tasks and
  \gls{rr} trails discrete acquisition optimization on the tabular tasks by a small median normalized
  regret of 0.007. In contrast, using \gls{gp}-\gls{bo} with a global surrogate (the variants and HEBO)
  instead of the other baselines is the first-order effect (\cref{sec:quality}).
\item \textbf{Source of the Delay:} We show that the delay of \gls{rr} comes from the steps in which
  the relaxed acquisition peaks inside an evaluated cell. Ax's native deduplication does not recover
  these steps, and a replacement by the best unobserved neighbouring grid point recovers them only in
  a loop that rounds the relaxed maximizer alone (\cref{sec:quality}, Appendix~\ref{app:loop}).
\item \textbf{Drivers of Wasted Evaluations:} We show that \gls{rr} re-proposes up to 63\% of its
  trials under an exploitative acquisition and more than half of its trainings on the coarsest CIFAR-10
  encodings, and that the waste grows with the grid step, exploitative acquisition functions, a
  noise-free \gls{gp}, a small initial design and batching. By intervening on the \gls{gp}
  lengthscale, we show that it modulates the waste but does not remove it, since on coarse grids both
  short and long lengthscales re-propose evaluated points. We also identify a second failure mode in
  mixed spaces, i.e., duplicates at boundary optima, which also affects discrete acquisition
  optimization (\cref{sec:waste,sec:mechanism}).
\item \textbf{CIFAR-10 Case Study:} We show that, with a validation split, test-only final
  evaluation and five runs per method, every method ends within 0.35~pp of test accuracy on our two
  search spaces and the best \gls{bo} variant beats quasi-random search by 0.29~pp (10 of 10 paired
  runs, Holm--Bonferroni-adjusted $p=0.014$), which is about the size of training-seed noise
  (\cref{sec:cifar}).\footnote{Code, configurations and all run logs will be released for
  reproducibility purposes.}
\end{enumerate}

\section{Related Work}
\label{sec:related}

\noindent\textbf{Hyperparameter Optimization of Neural Networks.}
\gls{hpo} methods range from model-free search, such as random and quasi-random search, to
model-based optimizers that fit a surrogate of the objective: the \acrlong{tpe} (\acrshort{tpe}, \citealp{bergstra2011algorithms})\glsunset{tpe}, \acrlong{smac} (\acrshort{smac}, \citealp{hutter2011sequential, lindauer2022smac3})\glsunset{smac} with random forests, and
\gls{gp}-based \gls{bo} \citep{snoek2012practical, shahriari2016taking}, including trust-region variants
such as TuRBO \citep{eriksson2019scalable}. They are implemented in Optuna
\citep{akiba2019optuna}, SMAC3 and Ax \citep{olson2025ax} and reviewed by \citet{bischl2023hyperparameter}. In the black-box optimization challenge analyzed by \citet{turner2021bayesian}, \gls{bo} clearly outperformed random search on machine learning tuning problems, and its winner, HEBO \citep{cowenrivers2022hebo}, relies on input warping and a multi-objective
acquisition ensemble. However, all these comparisons take the integer treatment of each library as
given and do not ablate it.

\noindent\textbf{Integer Inputs in \acrshort{gp}-Based \acrshort{bo}.}
\citet{garridomerchan2020dealing} showed that relaxing integers ignores the fact that the
objective is constant on each rounding cell, and proposed to round inside the kernel, which makes
the \gls{gp} piecewise constant. \citet{luong2019bayesian} instead adjust the kernel lengthscale with
respect to the grid spacing, together with the exploration term of the acquisition, so that rounded proposals do not collapse onto observed points.
We do not reimplement their method. However, in our setting, fixing or bounding the lengthscale does not prevent duplicates on coarse grids (\cref{sec:mechanism}).
\citet{daulton2022bayesian} optimize the expectation of the acquisition under a probabilistic
reparameterization of the discrete inputs, thus keeping gradient-based acquisition optimization
while scoring only grid points. Alternatively, the acquisition can be optimized directly over the
grid, exhaustively when the grid is small or by local search otherwise, and the BoTorch library \citep{balandat2020botorch} implements
both together with the exclusion of observed points. Closest to our
findings on final quality, \citet{karlsson2021continuous} report that surrogates over a continuous
relaxation perform well on expensive discrete problems. However, each remedy has been evaluated against its own baselines and surrogate, and no study compares them side by side. Finally, none of these works measures how much of the budget is spent on configurations that were already evaluated.

\noindent\textbf{Combinatorial and Mixed-Variable \acrshort{bo}.}
BOCS \citep{baptista2018bayesian} and COMBO \citep{oh2019combinatorial} model combinatorial
structure with dedicated surrogates, \acrlong{cocabo} (\acrshort{cocabo},
\citealp{ru2020bayesian})\glsunset{cocabo} combines multi-armed bandits
over categorical inputs with a \gls{gp} over continuous ones, and CASMOPOLITAN \citep{wan2021think}
uses trust regions in high-dimensional categorical and mixed spaces. Bounce
\citep{papenmeier2023bounce} embeds mixed spaces into nested subspaces of increasing dimension;
\citet{dreczkowski2023framework} provide a common framework and benchmark for these methods.
However, these works target categorical structure and large discrete spaces. Our focus is narrower and
practical: ordinal integer hyperparameters inside a standard \gls{gp}-\gls{bo} pipeline. To the best of our
knowledge, no prior study holds the surrogate and the acquisition function fixed while varying
only the integer treatment, and measures the resulting waste across grid steps, acquisition
settings and real \gls{hpo} tasks.

\noindent\textbf{Benchmarking \acrshort{hpo}.}
We compare optimizers over many tasks with Friedman/Iman--Davenport tests and Nemenyi \glspl{cd} on mean ranks \citep{demsar2006statistical, iman1980approximations}, paired
Wilcoxon tests with Holm--Bonferroni correction \citep{holm1979simple}, Bayesian signed-rank tests
\citep{benavoli2017time} and \gls{bca} bootstrap intervals
\citep{efron1987better}. Following \citet{bouthillier2021accounting}, we estimate the training
noise of the neural network case explicitly.
\section{Preliminary Notions}
\label{sec:prelim}

\subsection{Bayesian Optimization with Gaussian Processes}
\label{sec:bo}
We consider the maximization of a black-box objective $f:\mathcal{X}\to\mathbb{R}$, e.g., the
validation accuracy of a network trained with hyperparameters $x$, and handle minimization
problems by negating $f$. Each evaluation returns a noisy observation $y=f(x)+\varepsilon$ with
$\varepsilon\sim\mathcal{N}(0,\sigma_\varepsilon^2)$. \gls{bo} places a \gls{gp} prior on $f$, specified by a covariance function, or kernel,
$k(x,x')$, which measures the similarity between two configurations and thus determines how
much an observation at $x'$ informs the posterior at $x$. After $n$ evaluations
$\mathcal{D}_n=\{(x_i,y_i)\}_{i=1}^{n}$, the posterior of $f(x)$ is Gaussian with
mean $\mu_n(x)$ and standard deviation $\sigma_n(x)$, both available in closed form
\citep{garnett2023bayesian}. We use the Mat\'ern-5/2 kernel with \gls{ard},
\begin{equation}
k(x,x') = \sigma_f^2\left(1+\sqrt{5}\,r+\tfrac{5}{3}r^2\right)e^{-\sqrt{5}\,r},
\qquad r=\Big(\textstyle\sum_{j=1}^{d}(x_j-x'_j)^2/\ell_j^2\Big)^{1/2},
\label{eq:kernel}
\end{equation}
where the lengthscale $\ell_j$ sets the distance along dimension $j$ over which an observation
informs the posterior. The next configuration maximizes an acquisition function $\alpha_n$ that
trades off a high mean against a high uncertainty. One of those is \gls{ei} defined as 
$\mathrm{EI}_n(x)=\mathbb{E}[\max(f(x)-f_n^\ast,0)\mid\mathcal{D}_n]$ for the incumbent value
$f_n^\ast$, while the \acrlong{qlognei} (\acrshort{qlognei}, \citealp{ament2023unexpected})\glsunset{qlognei}
optimizes the logarithm of a batch version of \gls{ei} that integrates over the uncertainty of the
incumbent and is numerically easier to maximize. Another common acquisition function is the \gls{ucb} defined as
$\mathrm{UCB}_n(x)=\mu_n(x)+\sqrt{\beta}\,\sigma_n(x)$, where a small $\beta$ favors exploitation.
In standard implementations, $\alpha_n$ is maximized over a continuous box by multi-start
gradient ascent \citep{wilson2018maximizing, balandat2020botorch}, after an initial space-filling
design of $n_0$ points.

\subsection{Integer Search Spaces}
Let $x\in\mathcal{X}=\prod_{j=1}^{d}\{0,\dots,n_j-1\}$ index a grid, where each index maps to a
physical value $v_j(x_j)$, e.g., a width $v_j(x_j)=K\,(x_j+1)$ for a grid step $K$, a power of
two, or an entry of a lookup table. We call $K$ the \emph{granularity} of the grid: for a fixed
physical range, a larger $K$ means fewer and coarser levels.

\begin{figure}[t]
\centering
\includegraphics{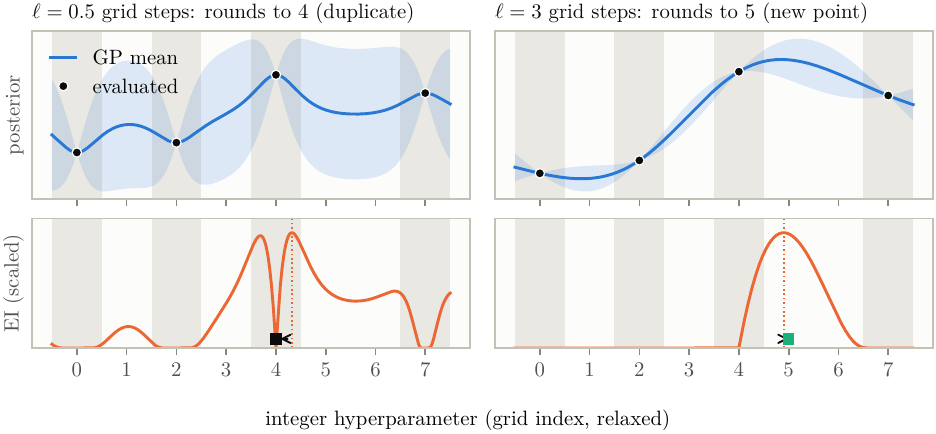}\vspace{-10pt}
\caption{\textbf{Relax-and-Round.} Illustration of \acrshort{rr} on a one-dimensional grid of 8 levels with four evaluated points
(black dots) and their rounding cells (shaded). \textbf{\emph{Top:}} \acrshort{gp} posterior mean and $\pm2\sigma_n$ band. \textbf{\emph{Bottom:}}
\acrshort{ei} over the relaxed interval and its maximizer (dotted line), and the arrow from the maximizer to the grid point it is rounded to (square): a black square is an evaluated point (a duplicate), a green square a new grid point. With a lengthscale of 0.5 grid steps (\textbf{\emph{Left}}), the posterior variance recovers inside
the cell of the best point, \acrshort{ei} peaks next to it, and rounding returns the evaluated point. With 3
grid steps (\textbf{\emph{Right}}), the cell is already explained by its observation and \acrshort{ei} peaks at a new grid
point.\vspace{-10pt}}
\label{fig:rr}
\end{figure}

\subsection{Relax-and-Round}
\gls{rr} fits a \gls{gp} on the relaxation $\prod_j[-\tfrac12, n_j-\tfrac12]$, maximizes $\alpha_n$ over that
box, and evaluates $\mathrm{round}(\arg\max\alpha_n)$. Each grid point thus owns a rounding cell
of unit width, and every continuous proposal inside the cell of an evaluated point is a
\emph{duplicate}. \Cref{fig:rr} illustrates why duplicates occur: whether the acquisition
maximizer leaves the cell of an evaluated point depends on how far the posterior variance
recovers inside that cell, which depends on the lengthscale measured in grid steps.

\subsection{Waste, Delay and Budget Views}
A \emph{duplicate} is a \gls{bo}-phase proposal whose canonical grid configuration was already
evaluated in the same run, and a method that makes it \emph{re-proposes} that configuration. We
measure two distinct costs of integer handling. The \emph{waste} is the \emph{duplicate rate}, i.e.,
the fraction of duplicates among the \gls{bo}-phase proposals within the budget $B$. The \emph{delay}
is measured by the number of evaluations, duplicates included, until the incumbent reaches a
normalized regret of 0.01 (\cref{sec:setup}), set to $B+1$ when it never does. Since duplicates may be free in practice (e.g., with a cache), we report every
budget-dependent metric in two views: \emph{counted}, where duplicates consume budget, and
\emph{unique}, where only distinct configurations count.

\section{Integer-Handling Methods}
\label{sec:methods}

\subsection{Shared Components}
All model-based variants share the initial design of their (task, seed) pair, i.e., the first
$n_0=2d+1$ points of a scrambled Sobol sequence \citep{owen1998scrambling}, attached to every
optimizer including the baselines. All variants except Ax-default share one surrogate: a \texttt{SingleTaskGP}
with the \gls{ard} Mat\'ern-5/2 kernel of \cref{eq:kernel}, BoTorch's dimension-scaled lengthscale
prior \citep{hvarfner2024vanilla} and inferred observation noise. All of them use \gls{qlognei}. The variants run on Ax~1.3.1 and BoTorch~0.18.1,
except \mPR and \mRRnb, which run in a pure-BoTorch loop with the same surrogate but without Ax's
winsorization of the outcomes (Appendix~\ref{app:settings}).

\subsection{Integer-Handling Variants}
The variants differ only in how integers enter the \gls{bo} loop, and they intervene at different
stages (\cref{tab:methods}). \mRRAx and \mRRext relax the integers, the former inside Ax and the latter
outside the model. \mRRcache and \mRRdedup act after the proposal, by serving duplicates from a cache or by
rejecting them. \mKRound changes the model, by rounding inside the kernel. \mPR and \mDisc change the
acquisition optimization, so that only grid points are scored. \mAxDef, which a user of current Ax
obtains without any configuration, combines discrete acquisition optimization with the exclusion of
observed points. Two controls isolate the remaining factors: \mDisc without exclusion and, in a
pure-BoTorch loop, \mRRnb.

\begin{table}[t]
\caption{Integer-handling variants and the stage of the \acrshort{bo} loop at which each one handles
integers.}
\label{tab:methods}
\centering
\small
\begin{tabular}{llp{0.68\textwidth}}
\toprule
& Stage & Description \\
\midrule
\mRRAx & none (\acrshort{rr}) & \acrshort{rr} as in Ax $<$1.0: integers relaxed by Ax's legacy \texttt{IntToFloat}
  transform, acquisition optimum rounded, no deduplication. \\
\mRRext & none (\acrshort{rr}) & \acrshort{rr} outside the model: float parameters on $[-\tfrac12, n-\tfrac12]$, rounded
  inside the objective, so the \acrshort{gp} sees the unrounded inputs. \\
\mRRcache & proposal & \mRRAx with a cache: a re-proposed configuration is not re-evaluated, its stored
  value is returned. \\
\mRRdedup & proposal & \mRRAx with Ax's native deduplication (duplicates are rejected and, up to five times,
  regenerated). \\
\mKRound & model & Rounding inside the kernel \citep{garridomerchan2020dealing}: \mRRext with BoTorch's
  \texttt{Round} input transform in its exact (non-approximate) form. \\
\mPR & acquisition & Probabilistic reparameterization \citep{daulton2022bayesian}: Monte Carlo expectation of the
  acquisition over the rounding distribution, then the best of samples from the optimized one. \\
\mDisc & acquisition & Discrete acquisition optimization: integers kept discrete, exhaustive
  evaluation of $\alpha_n$ on the grid up to $2\times10^6$ points with observed points excluded; beyond that size and in mixed spaces, BoTorch's mixed alternating optimizer, which does not exclude them. \\
\mAxDef & acquisition & The current Ax default (\texttt{choose\_generation\_strategy}): discrete
  integers, enumeration up to $10^4$ points or local search, observed points excluded, Ax's
  default surrogate. \\
\midrule
\mDisc, no excl. & acquisition & \mDisc without the exclusion of observed points (Tier A, $d\le4$). \\
\mRRnb & proposal & \acrshort{rr} in a pure-BoTorch loop; a duplicate is replaced by the unobserved
  grid point with the highest acquisition among its neighbours (Tier A, $d=2$). \\
\bottomrule
\end{tabular}
\vspace{-10pt}
\end{table}

\mRRAx reproduces the behavior of Ax versions before 1.0, whose default transforms relaxed integers. As shown in \cref{sec:waste}, it stalls on evaluated configurations. Since \mAxDef
differs from \mDisc in more than the integer treatment (surrogate defaults and acquisition
optimizer), we treat it as a reference point. \mPR is our own implementation of the Monte Carlo estimator of \citet{daulton2022bayesian} with the shared acquisition (Appendix~\ref{app:settings}).

\noindent\textbf{Baselines.}
The seven baselines cover model-free search (random search and Sobol), model-based \gls{hpo} with
non-\gls{gp} surrogates (\gls{tpe} and SMAC3), trust-region \gls{gp}-\gls{bo} (TuRBO-1, \citealp{eriksson2019scalable}),
\gls{gp}-\gls{bo} with input warping and an acquisition ensemble that rounds integers and drops observed
points (HEBO, \citealp{cowenrivers2022hebo}) and mixed-variable \gls{bo} (\gls{cocabo}, \citealp{ru2020bayesian}). Each of them receives the same initial design as the variants.

\section{Experimental Setup}
\label{sec:setup}

\begin{table}[t]
\caption{Benchmarks. $d$ is the number of hyperparameters, and the budget counts distinct
configurations on Tiers A and B and every training on Tier C.}
\label{tab:tasks}
\centering
\small
\begin{tabular}{@{}llllrr@{}}
\toprule
Tier & Benchmark (tasks) & Search space & $d$ & Budget & Seeds \\
\midrule
A & 4 synthetic functions $\times$ 3 noise levels (36) & 16 integer levels per dim & 2, 4, 8 & $\max(20d,50)$ & 30 \\
B & LCBench via YAHPO Gym (15) & 3 integers, 4 floats & 7 & 100 & 20 \\
B & HPOBench tabular \acrshort{mlp} (5) & 5 ordinal indices & 5 & 100 & 20 \\
B & JAHS-Bench-201 (3) & mixed, 8 categorical & 13 & 100 & 20 \\
C & CIFAR-10 \acrshort{cnn} (2 search spaces) & 4 widths, 3 learning rates & 5 & 50 & 5 \\
\bottomrule
\end{tabular}
\vspace{-10pt}
\end{table}

\subsection{Benchmarks}
\Cref{tab:tasks} summarizes the three tiers. \textbf{Tier A} contains Rosenbrock, Ackley, Levy
and Styblinski--Tang in $d\in\{2,4,8\}$, with each dimension discretized to 16 levels on a grid
aligned so that the continuous minimizer is a grid point, i.e., the grid optimum is known exactly.
Additive Gaussian noise has a standard deviation of 0, 1.8\% or 6\% of the gap between the median
and the optimum. \textbf{Tier B} contains 15 LCBench tasks \citep{zimmer2021autopytorch} through
the YAHPO Gym surrogate \citep{pfisterer2022yahpo}, with three integers of 5, 961 and 497 values
and four floats (five tasks chosen first and ten drawn at random with a fixed seed), and 5
HPOBench tabular \gls{mlp} tasks \citep{eggensperger2021hpobench}, whose five
ordinal hyperparameters are encoded as integer indices over 30{,}000 grid points.
JAHS-Bench-201 \citep{bansal2022jahs} is in Appendix~\ref{app:jahs}. \textbf{Tier C} (\cref{sec:cifar}) is
a \gls{cnn} with integer layer widths on CIFAR-10 \citep{krizhevsky2009learning}.

\subsection{Protocol and Metrics}
\noindent\textbf{Granularity Study.}
To isolate the grid step, we fix the physical range and represent a dimension of $L=192$ nominal
levels with step $K$ and $L/K$ levels, $K\in\{1,2,4,8,12,16\}$, on Tier A with $d\in\{2,4\}$ and
20 seeds, and similarly for the LCBench width and batch size, including a power-of-two encoding.

\noindent\textbf{Budgets and Metrics.}
On Tiers A and B, a run continues until it has evaluated the budget of distinct configurations,
capped at twice the budget, which enables both budget views, while Tier C counts every training.
The incumbent is the configuration with the best \emph{observed} value, and its regret is
measured on the noiseless objective where available. Normalized regret divides by the gap
between the optimum and a reference value. For Tier A, the reference is the median of the
function over 4{,}096 random grid points. For Tier B, the best value seen in any run replaces the optimum and the
reference is the median of the initial design. Differences between rates and between accuracies
are absolute, in pp.

\subsection{Statistical Analysis}
For each task, we compare methods with paired Wilcoxon tests over seeds and Holm--Bonferroni correction, and $p_\mathrm{HB}$ denotes an adjusted $p$-value.
Across tasks, we average over seeds, rank the methods on each task, and apply the
Friedman/Iman--Davenport test followed by Nemenyi \glspl{cd} and Holm--Bonferroni-corrected Wilcoxon tests over
tasks. We report \gls{bca} 95\% intervals and Bayesian signed-rank tests with a \gls{rope} of 0.01 in normalized regret, which equals the target regret of \cref{sec:quality} (\cref{tab:rope} reports 0.005 and 0.02).

\section{Experiments}
\label{sec:experiments}

\noindent\textbf{Objectives.}
The evaluation focuses on three research questions: \one\ Does the integer treatment change the
regret that \gls{bo} reaches, at the end and along the search? \two\ How much of the budget does
\gls{rr} waste, and which settings drive the waste? \three\ What causes the duplicates, i.e., the
grid, the lengthscale or the acquisition optimizer? Further statistics for each question are
reported in Appendix~\ref{app:extra}.

\subsection{Final Quality}
\label{sec:quality}

\begin{table}[t]
\caption{Mean rank of final normalized regret (counted view, lower is better, number of tasks
in parentheses) and mean duplicate rate of the \acrshort{bo} phase. Methods whose ranks differ by less than
the \acrshort{cd} are not significantly different (Nemenyi, $\alpha=0.05$). \mRRcache
re-proposes as often as \mRRAx, but serves its duplicates from the cache, so they cost no
evaluation time while still consuming trials in the counted view.}
\label{tab:main}
\centering
\small
\input{tables/main.tex}
\vspace{-10pt}
\end{table}

\begin{figure}[t]
\centering
\includegraphics{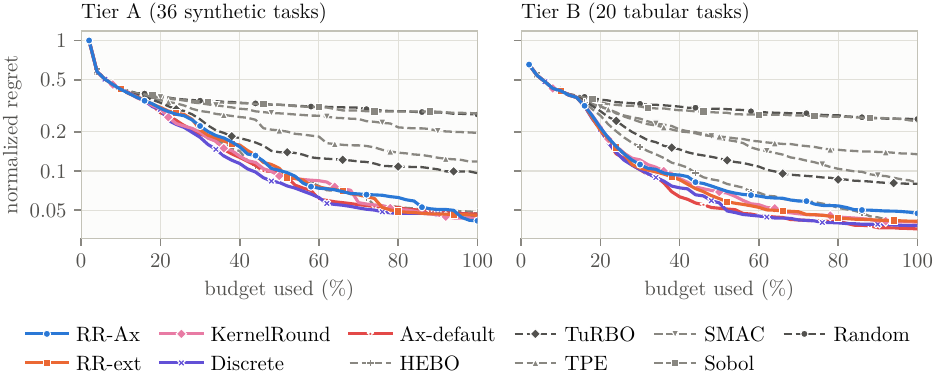}\vspace{-10pt}
\caption{\textbf{Anytime Regret.} Normalized regret of the incumbent against the share of the budget used (duplicates
counted; median over seeds per task, then mean over tasks, log scale). The integer-handling
variants (solid) form one bundle well below the baselines (dashed) from about a quarter of the
budget on, and HEBO joins it by the end. Tier A budgets depend on $d$, so the curves are aligned on the budget share.\vspace{-10pt}}
\label{fig:curves}
\end{figure}

\Cref{tab:main} reports the mean rank of final normalized regret, \cref{fig:curves} the regret along
the search and \cref{fig:cd} the \gls{cd} diagrams. Three main observations emerge.

\noindent\textbf{\one\ The integer treatment barely changes final regret.}
On Tier A, all eight variants fall within one \gls{cd} (3.57) of each other, and no pairwise
comparison between them survives Holm--Bonferroni correction. The Bayesian signed-rank test
also places 99.2\% of the posterior mass of \mRRAx vs.\ \mAxDef inside the \gls{rope}. Tier B separates the
variants more: \mDisc, \mAxDef and \mRRext lead (\cref{tab:main}), and \mAxDef and \mRRext improve on \mRRAx
($p_\mathrm{HB}=0.023$ and $0.008$). However, the median per-task gap between \mRRAx and \mDisc is only
0.007, so the posterior mass of \mRRAx vs.\ \mAxDef splits between the \gls{rope} (0.45) and \mAxDef (0.55), and
moves to \mAxDef (0.90) only with a \gls{rope} of 0.005 (\cref{tab:rope}). HEBO, which also relaxes and
rounds but drops observed points, ties with every variant on both tiers.

\noindent\textbf{\two\ Using \gls{gp}-\gls{bo} has a much larger effect than its integer treatment.}
On Tier A, every variant is significantly better than \gls{tpe}, TuRBO and \gls{smac} (all
$p_\mathrm{HB}<0.05$), which in turn rank ahead of random search, Sobol and \gls{cocabo}. On Tier B,
the variants and HEBO form one Nemenyi group. The best of them (\mDisc, \mAxDef, \mRRext, \mPR and HEBO) are
separated from TuRBO, \gls{smac} and \gls{tpe} (mean ranks 9.95--10.80), and every variant is
separated from random search and Sobol (14.00--14.90). As shown in \cref{fig:curves}, this ordering
already holds during the search: with half of the budget, every variant reaches a lower regret on
Tier B than TuRBO and \gls{smac} with their full budget.

\noindent\textbf{\three\ \gls{rr} delays the search when the relaxed acquisition peaks inside an evaluated cell.}
\mRRAx needs about four more evaluations than \mDisc on Tier A (\cref{tab:delay},
$p_\mathrm{HB}=0.005$), and Ax's deduplication does not help, since \mRRdedup is never faster. At $d=2$, the relaxed maximizer falls inside the cell of an evaluated point in about half
of the steps, and Ax then either re-evaluates the point or falls back on one of low acquisition
(Appendix~\ref{app:stats}). Duplicates alone do not cause the delay, since \mDisc without exclusion is
as fast as \mDisc. In a BoTorch loop at $d=2$, rounding every restart and keeping the best rounded point, as Ax does,
alone cuts the duplicates of \gls{rr} from 65\% to 21\% (Appendix~\ref{app:loop}). With this selection, \gls{rr} stays about three evaluations slower than enumeration (\cref{tab:delay},
bottom; posterior probability 0.96, $p_\mathrm{HB}=0.28$), and replacing duplicates with the best
unobserved neighbour no longer helps, since it speeds up only the plain loop. Together with the
comparison in Ax, this indicates that the delay comes from the relaxation itself, and optimizing the
acquisition on the grid avoids it.

\subsection{Wasted Evaluations}
\label{sec:waste}
\label{sec:ablation}

\begin{table}[t]
\caption{Ablations on Tier A ($d=4$, 16 levels, 4 functions $\times$ 2 noise levels, 20 seeds).
Each row changes one setting of the base variant. $\Delta$ is the variant minus the base,
$p_\mathrm{HB}$ is from paired Wilcoxon tests, a positive $\Delta$ regret is worse, and changes with
$p_\mathrm{HB}<0.05$, with the corresponding duplicate rates, are in \textbf{bold}. $^\dagger$\mDisc
with \acrshort{ucb} $\beta=0.5$ and without the exclusion of observed points, compared with \mDisc by
a separate paired test (outside the Holm--Bonferroni family).}
\label{tab:ablation}
\centering
\small
\input{tables/ablation.tex}
\vspace{-10pt}
\end{table}

\begin{wraptable}{r}{0.37\textwidth}
\vspace{-50pt}
\caption{\textbf{Delay.} Mean evaluations to the target. Bottom: BoTorch loop, plain or with Ax's
settings (Appendix~\ref{app:loop}).}
\label{tab:delay}
\centering
\footnotesize
\input{tables/delay.tex}
\vspace{-10pt}
\end{wraptable}
The duplicate rates of \cref{tab:main} (right) show how many proposals each method wastes, and
\cref{tab:ablation} changes one optimizer setting at a time. We draw three observations.

\noindent\textbf{\one\ Only methods that exclude observed points never re-propose.}
On Tier A, \mRRAx and \mRRext re-propose 11--13\% of their \gls{bo} trials and TuRBO 45\%, since its trust
region shrinks below one grid step and keeps rounding to the same point. \mPR re-proposes 4.2\%,
since it samples its proposal without excluding observed points. In contrast, \mRRdedup, \mDisc, \mAxDef and HEBO
never re-propose a point.

\noindent\textbf{\two\ Duplicates make \gls{rr} stall but rarely worsen final regret.}
On the noise-free Tier A tasks, most runs of \mRRAx propose some configuration at least five times
(Appendix~\ref{app:stats}). Nevertheless, its final regret is close to that of \mAxDef (0.0086 vs.\
0.0075). A noise-free \gls{gp} is the exception, since it raises the waste further and also worsens
final regret ($p_\mathrm{HB}=0.045$).

\noindent\textbf{\three\ Exploitation, a noise-free \gls{gp}, small initial designs and batches increase the waste.}
Exploitation is the strongest driver and the only one that worsens final regret
(\cref{tab:ablation}): \gls{ucb} with $\beta=0.5$ makes \mRRAx re-propose most of its trials, while
$\beta=8$ is harmless. Most of this waste comes from the acquisition and not from rounding, since
under the same \gls{ucb} \mDisc without exclusion re-proposes only 4.3~pp less than \mRRAx ($p<0.001$)
and also worsens final regret. The analytic \gls{ei} raises the waste as well, consistent with the
optimization difficulties of \gls{ei} shown by \citet{ament2023unexpected}, and so does a
noise-free \gls{gp}, since on a noise-free objective a re-evaluated point changes
neither the posterior nor the next proposal. A smaller initial design and batches also increase
the waste, and in batches only the methods that deduplicate, \mRRdedup and \mAxDef, avoid it
(Appendix~\ref{app:stats}). Excluding observed points, \mDisc is immune to the noise model, \gls{ei}
and the kernel.

\subsection{Causes of the Duplicates}
\label{sec:mechanism}
\label{sec:boundary}

\begin{figure}[t]
\centering
\includegraphics{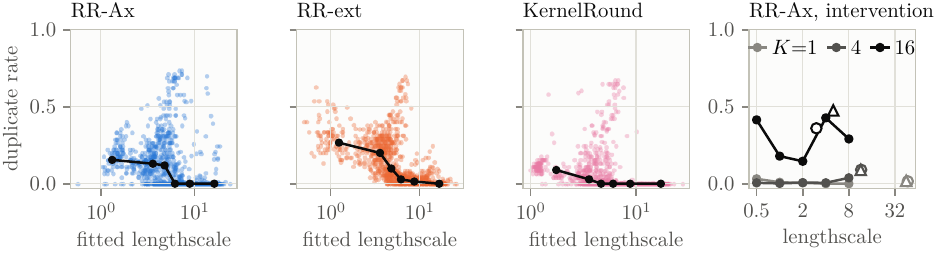}\vspace{-10pt}
\caption{\textbf{Duplicates and Lengthscale.} \textbf{\emph{Left:}} duplicate rate of each run against its median
fitted \acrshort{gp} lengthscale in grid steps (Tier A, medians of six quantile bins in black).
\textbf{\emph{Right:}} intervention on \mRRAx ($d=2$, mean over 4 functions and 20 seeds per grid step $K$), with the
lengthscale fixed at 0.5 to 8 grid steps (lines), fitted (circles), and fitted above a floor of 5
grid steps (triangles).\vspace{-10pt}}
\label{fig:ls}
\end{figure}

\Cref{fig:dupk} relates the duplicate rate to the grid step, \cref{fig:ls} to the lengthscale, and
Tier B adds mixed spaces. Three findings stand out.

\noindent\textbf{\one\ Duplicates grow with the grid step, partly because of rounding.}
At $K=1$ (192 levels), the relaxed variants re-propose at most 7\% of their proposals, whereas at
$K=16$ they re-propose up to half of them (\cref{fig:dupk}). \mDisc without exclusion stays close
behind, so on a small, coarse grid the acquisition maximum itself is often an evaluated point, and
rounding adds from 0.03~pp at $K=1$ to 7.2~pp at $K=16$. The two parts grow with different quantities
(\cref{fig:dupk}, right). The duplicates of \mDisc without exclusion follow the share of the grid
that has been evaluated, and its $d=4$ points fall on its $d=2$ curve. In contrast, at the same
share, the $d=4$ runs use a much coarser grid ($K=12$ and $16$ vs.\ $K\le2$), and \mRRAx re-proposes more than
twice as often as at $d=2$, since the duplicates added by rounding track the grid step. At a fixed
grid step, both parts weaken with the dimension, since a run evaluates a smaller share of a larger
grid (\cref{fig:dupk_full}).

\noindent\textbf{\two\ The lengthscale modulates the waste but does not remove it.}
In \cref{fig:ls} (left), the median duplicate rate of \mRRAx is 12--16\% for fitted lengthscales below
5.4 grid steps and zero above, as \cref{fig:rr} predicts. However, the fitted
lengthscale is about a quarter of the range at every $K$, so in grid steps it mostly reflects the
resolution. When we set it (right), the rate at $K=16$ ranges from 15\% to 43\% without a monotone
trend, and a floor of 5 grid steps even raises it from 36\% to 48\% ($p_\mathrm{HB}<0.001$) and
worsens final regret by 0.029 ($p_\mathrm{HB}=0.011$), since a smooth posterior keeps the
acquisition maximizer inside the cell of the incumbent. Hence, only excluding observed points removes
duplicates at every grid step.

\noindent\textbf{\three\ Mixed spaces add a second failure mode at boundary optima.}
On HPOBench, whose hyperparameters are all integer indices, the grid effect appears as on Tier A
(\mRRAx 4--40\%, TuRBO 46--52\%, \mDisc and \mAxDef none). On LCBench, \mRRAx re-proposes 14--61\% of its trials
on six of the 15 tasks and \mDisc 14--58\% on five of them, although an exact duplicate requires all four
continuous hyperparameters to coincide. In every such duplicate, these hyperparameters sit at a
bound of the box, so the gradient-based optimizer returns to the same boundary point and rounding
plays no role. \mDisc is affected because, in mixed spaces, it uses BoTorch's alternating optimizer,
which does not exclude observed points, and only \mAxDef avoids this failure mode (at most 0.4\%).
These boundary optima belong to the YAHPO surrogate, which may extrapolate poorly at the edges of
the box, so they need not exist in the real training curves. However, the duplicates only require the
acquisition maximum to sit at a bound. Hence, any loop that optimizes continuous hyperparameters by gradient needs to exclude observed points, even with discrete integers.

\section{Case Study: CIFAR-10 \acrshort{cnn}}
\label{sec:cifar}

\subsection{Protocol}
The \gls{cnn} has four $5\times5$ convolutions with batch normalization and one max-pooling layer.
The first search space has four widths $12j$, $j\in\{1,\dots,16\}$ (step $K=12$, widths up
to 192) and a learning rate in $\{10^{-2},10^{-3},10^{-4}\}$, and a second space doubles the
width bound to 384 to check that the optimum found is not an artifact of the bound. The protocol
separates method differences from training noise in four ways:
\begin{enumerate}
\item a fixed stratified 45k/5k train/validation split, so that the optimizer only sees
  validation accuracy and the best epoch is also selected on validation;
\item the test set is used only for the final evaluation;
\item five independent \gls{bo} runs per method and space;
\item the observation noise is inferred by the \gls{gp} and not fixed to zero.
\end{enumerate}
To afford this, we train for 15 epochs instead of 25, since on 20 configurations the ranking
after 15 epochs closely agrees with full training (Kendall $\tau=0.95$, Spearman $\rho=0.99$, 4
of the top 5 shared). Each run has 50 trainings, including the 11-point Sobol design. At the end
of each run, we retrain its three best observed configurations with five fresh seeds, select the
one with the highest mean validation accuracy as the final configuration, and report its mean
test accuracy. Training used bf16 on one \gls{gpu} type and fp32 on another (15 of the 50 runs), with no significant effect on accuracy (Appendix~\ref{app:cifar}).

\noindent\textbf{Training Noise.}
Retraining four configurations of the first space with five seeds each, a hand-chosen one and
the three best of a single-seed \gls{bo} search, gives a pooled standard deviation of validation
accuracy of 0.35\%, while consecutive configurations among the five best of that search differ
by at most 0.19~pp. A single-seed ranking of the top configurations is not
identifiable, and a noise-free \gls{gp} is misspecified here.

\begin{table}[t]
\caption{CIFAR-10 validation and test accuracy (\%) of the final configuration (mean $\pm$ std
over 5 runs) and mean number of duplicate trainings out of 50. $\Delta$: mean difference from
Sobol over the 10 runs of both spaces, paired by space and seed (pairs won in parentheses). In
\textbf{bold}, results significantly better than Sobol (Wilcoxon, Holm--Bonferroni).}
\label{tab:cifar}
\centering
\small
\setlength{\tabcolsep}{3.4pt}
\input{tables/cifar.tex}
\vspace{-10pt}
\end{table}

\subsection{Results}
\label{sec:cifar_results}
\noindent\textbf{Accuracy.}
\Cref{tab:cifar} (and \cref{fig:cifar} in Appendix~\ref{app:cifar}, per run) shows that every
method, including Sobol and random search, ends within 0.35~pp of the others on test accuracy.
The advantage of the best \gls{bo} variant over Sobol is small but consistent: \mDisc beats Sobol in all
10 pairs ($p_\mathrm{HB}=0.014$ on both metrics). \mAxDef and \mRRAx point in the same direction but do not survive the correction ($p_\mathrm{HB}\geq0.06$), and no \gls{bo} variant significantly
beats random search.

\noindent\textbf{Wasted Trainings.}
\gls{rr} wastes trainings here too: \mRRAx re-proposes 14.6 of its 50 trainings on average in the first
space (between 0 and 23 per run) and 2.0 with widths up to 384. Coarser encodings make it worse,
as \mRRAx re-proposes 25--32 trainings with $K=24$ and 28--34 with power-of-two widths, i.e., more than
half of the budget (Appendix~\ref{app:cifar}). In contrast, \mDisc and \mAxDef never re-propose a training. A
duplicate training draws a new seed, but with a training noise of 0.35\% it rarely changes the
incumbent.

\section{Discussion}
\label{sec:discussion}

\noindent\textbf{Final Regret Robustness.}
Duplicates delay the search but rarely trap it: with inferred noise, a duplicate keeps
reducing the posterior variance and the optimizer eventually moves on. In the unique view, \mRRAx is
not significantly worse than \mAxDef on either tier (Tier B: $p_\mathrm{HB}=0.26$), so the cost of
\gls{rr} is budget, while its remedies cost at most 3.2\,s per proposal (Appendix~\ref{app:cost}).

\noindent\textbf{Replication versus Waste.}
With noisy observations, evaluating a configuration again can be a deliberate strategy that
sharpens the estimate of a promising point \citep{binois2019replication}. The duplicates we count,
however, come from rounding or from an acquisition that keeps returning to the best region, are most
frequent on noise-free objectives, where a second evaluation returns the same value, and rarely
change the incumbent. Their only advantage, on the coarsest CIFAR-10 encodings, is that new
training seeds favor lucky draws in the best observed value (Appendix~\ref{app:cifar}).

\section{Limitations and Future Work}
\label{sec:limitations}

While our findings hold across synthetic, tabular and neural network tasks, the real-world
tiers are modest, with 20 tabular tasks, three JAHS tasks and two CIFAR-10 search spaces with
five runs each, so on CIFAR-10 differences of about 0.2~pp between methods are at the
limit of what we can resolve (e.g., \mAxDef against Sobol).
The tabular tasks rely on surrogates or lookup tables. The \gls{cnn} is also small and its search space has only layer widths and a three-level learning rate, to which accuracy is fairly insensitive, so the small spread among methods may reflect a flat space. Furthermore, \mAxDef bundles several differences from \mDisc, so their
equivalence does not show that the integer treatment is the only relevant factor. Finally, the CIFAR-10 runs
use 15 training epochs, justified by the rank agreement with 25 epochs but not identical to full
training, and our BoTorch-loop experiments cover only the $d=2$ synthetic tasks. Natural next steps are extending the study to categorical, conditional and multi-fidelity settings and testing grid-aware replacements in higher dimensions, batches and mixed spaces.

\section{Conclusion}
\label{sec:conclusion}

In this work, we have studied when relaxing integer hyperparameters and rounding the acquisition
optimum hurts \gls{gp}-based \gls{bo}. To this end, we have \one\ compared eight integer-handling variants with
seven \gls{hpo} baselines on paired initial designs; \two\ varied the factors that drive the waste one at
a time; and \three\ tuned a CIFAR-10 \gls{cnn} with validation-based selection and repeated runs. This
way, we have separated the cost of \gls{rr}, which is budget, from its effect on final quality, which
is small. \gls{rr} delays the search, since the relaxed acquisition often peaks inside the cell of an
evaluated point, and wastes up to 63\% of the trials under an exploitative acquisition and more than
half of the trainings on our coarsest CIFAR-10 encodings, mostly because observed points are not
excluded, while rounding adds a share that grows with the grid step. On our CIFAR-10 search spaces,
the best \gls{bo} variant improves on quasi-random search by only 0.29~pp, about the training-seed
noise. Therefore, integers should be kept discrete in the acquisition optimizer and observed points
excluded, as current Ax does by default. Where a relaxation is unavoidable, comparing the restarts of
the acquisition optimizer after rounding removes most duplicates in our $d=2$ experiments but not the
delay; observed points should still be excluded, also within batches and at boundary optima; and
exploitative acquisitions, noise-free \glspl{gp} and coarse encodings should be avoided.

\bibliography{references}
\bibliographystyle{tmlr}

\clearpage
\appendix
\crefalias{section}{appendix}
\crefalias{subsection}{appendix}
\crefalias{subsubsection}{appendix}
\begin{center}
    {\LARGE Bayesian Optimization for Hyperparameters Tuning in Neural Networks\par}
    \vspace{0.4em}
    {\Large Supplementary Material\par}
\end{center}

\vspace{0.4em}
\begingroup
    \etocsetlocaltop{chapter}
    \etocsetnexttocdepth{subsection}
    \etocsettocstyle{\section*{Contents}}{}
    {\small\linespread{0.9}\selectfont\localtableofcontents*}
\endgroup
\vspace{1em}
\clearpage

\section{Additional Results}
\label{app:extra}

\subsection{Further Statistics}
\label{app:stats}
\begin{figure}[h]
\centering
\includegraphics{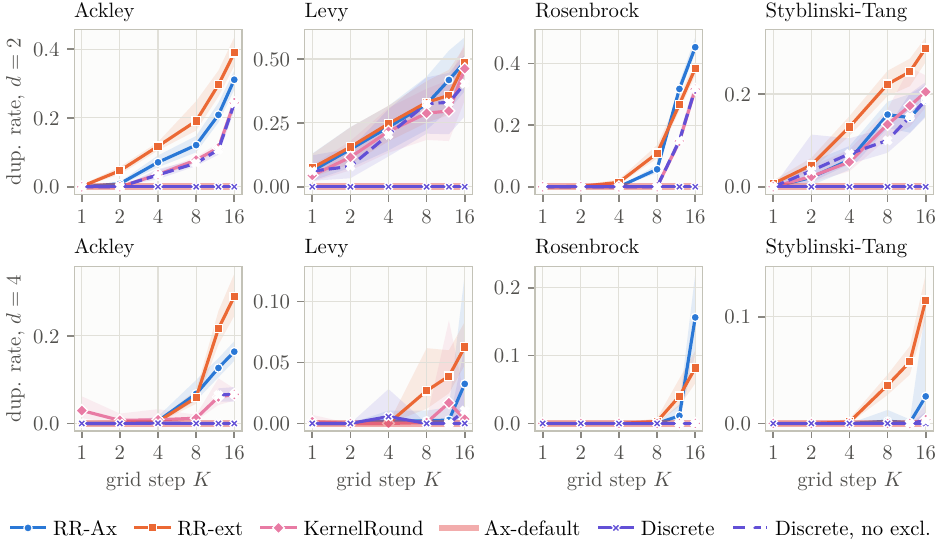}
\caption{\textbf{Grid Step for $d=2$ and $d=4$.} Duplicate rate against the grid step $K$ on Tier A, $d=2$ (\textbf{\emph{top}}) and $d=4$ (\textbf{\emph{bottom}}). At $d=4$, \mDisc without exclusion is run only at $K=12$ and $16$ (\cref{sec:mechanism}).}
\label{fig:dupk_full}
\end{figure}

\begin{figure}[h]
\centering
\includegraphics[width=\textwidth]{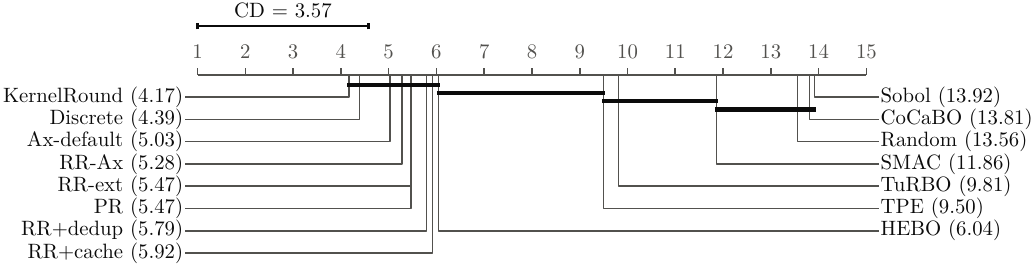}\\[2mm]
\includegraphics[width=\textwidth]{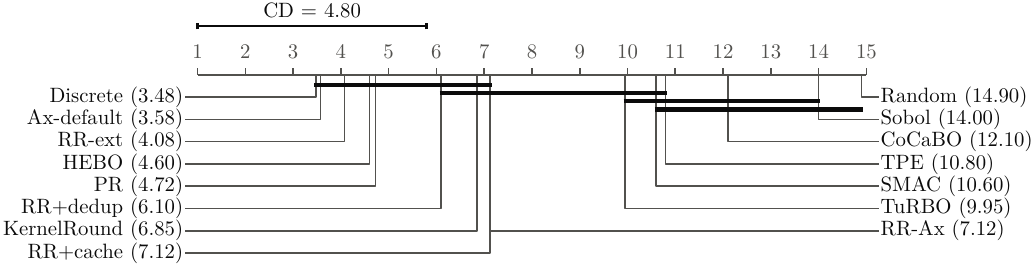}
\caption{\textbf{Critical Difference Diagrams.} \acrshort{cd} diagrams of final normalized regret (counted view) on Tier A (\textbf{\emph{top}}, 36 tasks) and
Tier B (\textbf{\emph{bottom}}, 20 tasks). Bars join methods that are not significantly different (Nemenyi,
$\alpha=0.05$).}
\label{fig:cd}
\end{figure}

We report here the statistics that complete \cref{sec:experiments}: the grid step per function
(\cref{fig:dupk_full}), the \gls{cd} diagrams (\cref{fig:cd}), the sensitivity to the \gls{rope}
(\cref{tab:rope}) and the following tests and rates.

\begin{table}[h]
\caption{Sensitivity of the Bayesian signed-rank test to the width of the \acrshort{rope} (final
normalized regret): posterior probability that a is better / that a and b are practically
equivalent / that b is better.}
\label{tab:rope}
\centering
\small
\input{tables/rope.tex}
\vspace{-10pt}
\end{table}

\begin{itemize}
\item \emph{Final quality.} On Tier A, the Friedman test rejects equal performance
  ($p=2\times10^{-64}$). In the unique view, \mKRound ranks first and \mAxDef fifth, and HEBO has mean rank 6.04
  against 4.17--5.92 for the variants. On Tier B, \mDisc narrowly misses an improvement over \mRRAx after
  correction over all 105 pairs ($p_\mathrm{HB}=0.060$), \mDisc and \mAxDef are practically equivalent
  (posterior mass in the \gls{rope} 1.00), and the first 10 tasks alone give the same ranking (mean
  rank of \mDisc 3.15) with a wider \gls{cd} (6.78 vs.\ 4.80).
\item \emph{Anytime quality.} With half of the budget, the variants reach 0.051--0.072 on Tier B,
  against 0.079 and 0.080 for TuRBO and \gls{smac} at the end, and 0.082--0.105 on Tier A, against
  0.096 for TuRBO. HEBO ends within the range of the variants (0.048 against 0.041--0.047 on Tier A,
  0.038 against 0.035--0.046 on Tier B) but trails them on Tier B at half of the budget (0.082), and
  \mRRAx lags on Tier B in the middle of the search (0.072 against 0.051 for \mAxDef). The area under the
  regret curve of \mRRAx is larger than that of \mDisc on both tiers ($p_\mathrm{HB}=0.035$ and $0.039$)
  and than that of \mAxDef on Tier B ($p_\mathrm{HB}=0.029$), with Holm--Bonferroni correction over the
  two tiers and the two references.
\item \emph{Delay decomposition.} We compare the mean number of evaluations to reach a normalized
  regret of 0.01 (\cref{tab:delay}) with Wilcoxon tests over tasks, Holm--Bonferroni-corrected over
  the six tests of the two tiers, and with Bayesian signed-rank tests with a \gls{rope} of one
  evaluation. \mRRAx is slower than \mDisc on 25 of 36 Tier A tasks ($p_\mathrm{HB}=0.005$) and on 11
  of 20 Tier B tasks ($p_\mathrm{HB}=0.12$, posterior probability that \mDisc is faster 0.97). \mRRdedup
  does not differ from \mRRAx on either tier ($p_\mathrm{HB}=0.13$): the posterior mass in the
  \gls{rope} is 0.71 and 0.99, and the probability that \mRRdedup is faster is 0.00 and 0.01. \mDisc is
  faster than \mRRdedup on 26 of 36 Tier A tasks ($p_\mathrm{HB}=0.002$) and on 11 of 20 Tier B tasks
  ($p_\mathrm{HB}=0.12$, posterior probability 0.96). On the 12 Tier A tasks with $d=2$, \mDisc without
  exclusion and \mDisc are practically equivalent (posterior mass in the \gls{rope} 1.00, equal means
  on 8 tasks).
\item \emph{Acquisition gap of \mRRdedup.} For the $d=2$ Tier A runs of \mRRdedup and \mDisc (10 seeds, 5{,}400
  proposals each), we refit the shared surrogate on the trials preceding each proposal (relaxed
  encoding for \mRRdedup, discrete for \mDisc) and evaluate \gls{qlognei} on every unobserved grid
  point. Maximized over the relaxed box with 20 restarts, the acquisition rounds onto an evaluated
  point in 47\% of the steps. Ax, however, rounds the end point of every restart and keeps the one
  with the highest acquisition after rounding, and this point is still an evaluated one in 23\% of
  all steps, close to the duplicate rate of \mRRAx at $d=2$ (25\%). In these steps, \mRRAx re-evaluates
  the point and \mRRdedup accepts a redraw of negligible acquisition value (median ratio of \gls{ei} to the
  best unobserved grid point below $10^{-10}$). In the other steps in which the relaxed maximizer
  falls inside an evaluated cell, the proposals of \mRRdedup reach a median ratio of 0.03, against 0.30
  when the relaxed maximizer is new and 0.70 for the proposals of \mDisc under the same refit. The
  ratio of \mDisc falls short of one since the refit differs from the fitted Ax model, so we compare
  the two methods rather than read the ratios as exact.
\item \emph{Stalls.} On the 8 noise-free Tier A tasks with $d\in\{2,4\}$, \mRRAx re-proposes 15.9\% of its
  trials, 59\% of its runs propose a single configuration at least five times, and one run proposes the
  same configuration 34 times. With a noise-free \gls{gp}, the rate rises to 19.3\% and the final
  normalized regret from 0.0086 to 0.024 ($p_\mathrm{HB}=0.045$).
\item \emph{Initial design and batches.} On the four noise-free ablation tasks, an initial design of
  $n_0=d+1$ points nearly doubles the duplicate rate of \mRRAx (6.5\%, against 3.5\% with $n_0=4d$),
  whereas the 2.6\% of \cref{tab:ablation} also averages over the noisy tasks. With batches of $q=4$
  on three LCBench tasks, the waste reappears inside the batch: \mRRAx re-proposes 9.9\%, \mDisc 8.9\%,
  \mRRext 6.2\%, \mRRdedup 0.1\% and \mAxDef none.
\item \emph{Optimizer settings.} A noise-free \gls{gp} adds 2.4~pp of duplicates
  ($p_\mathrm{HB}=0.024$) without a significant effect on regret ($p_\mathrm{HB}=0.63$), unlike in
  \cref{sec:waste}, since half of the eight ablation tasks are noisy. With $\beta=2$, \gls{ucb} still
  makes \mRRAx re-propose 41\%.
\item \emph{Lengthscale.} The median duplicate rate of \mRRext drops from 27\% to at most 3\% above 5.5
  grid steps, and that of \mKRound reaches zero above 4.2 steps. The fitted lengthscale of \mRRAx at $d=2$
  is 47, 12 and 3 grid steps at $K=1$, 4 and 16. TuRBO re-proposes at every lengthscale, since its trust
  region confines the search. With the lengthscale fixed, the rate stays at most 3\% at $K=1$ and,
  at $K=16$, is lowest at 2 steps (15\%) and highest at 4 steps (43\%). The floor of 5 steps raises
  the rate of \mRRext from 39\% to 57\%.
\item \emph{Mixed spaces.} On HPOBench, \mKRound stays at or below 3.4\%. On LCBench, the duplicates
  also have every integer at a bound on most tasks, and on LCBench 7593 \mRRAx re-proposes 33--45\% of
  its trials at every grid step $K$ and 63\% with power-of-two widths.
\end{itemize}

\subsection{BoTorch Loop and the \acrshort{rr} Baseline}
\label{app:loop}
To test whether a grid-aware replacement removes the delay of \gls{rr}, we run a pure-BoTorch loop
with the shared surrogate and \gls{qlognei} on the 12 Tier A tasks with $d=2$ (10 seeds). It runs
\gls{rr} (multi-start optimization over the box and rounding, no deduplication), \mRRnb, in which a
duplicate is replaced by the unobserved grid point with the highest acquisition within Chebyshev
radius $r=1,2,\dots$ of the rounded point (smallest $r$ with an unobserved point), and enumeration
of the acquisition on the unobserved grid points, each with plain settings and with those of Ax
(\cref{tab:loop}).

\begin{table}[h]
\caption{\textbf{BoTorch Loop.} Tier A, $d=2$, 12 tasks, 10 seeds (\mRRAx: the 30 seeds of the main
study). The settings of Ax are added to plain \acrshort{rr} one at a time: rounding the end point of
every restart and keeping the best rounded point, cells of width $1/n$ instead of the encoding
$i/(n-1)$, and winsorization. Dup.: duplicate rate of the \acrshort{bo} phase. Stalled: runs whose trials
31 to 50 contain at most two distinct configurations. Regret: final normalized regret.
Evaluations: to a normalized regret of 0.01 (counted view). Means over tasks.}
\label{tab:loop}
\centering
\small
\input{tables/loop.tex}
\end{table}

In its plain form, the loop rounds only the relaxed maximizer, which often lies inside the cell of
the incumbent, so \gls{rr} re-proposes 65\% of its trials, stalls in 57\% of the runs and ends at a
worse regret than enumeration ($p_\mathrm{HB}=0.029$), although, refit on the same trials, the two
loops give the same surrogate hyperparameters. Ax instead rounds the end point of every restart and
keeps the restart with the highest acquisition after rounding, and since an observed point has a
low acquisition, another restart usually wins. This selection alone removes most duplicates and
stalls, whereas Ax's encoding and its winsorization do not. With all three settings, the loop
reproduces the duplicates, stalls and final regret of \mRRAx (\gls{rope} mass 0.89 on regret), and it
remains 2.6 evaluations slower ($p=0.03$ uncorrected). The plain baseline therefore stalls because
of how it optimizes the relaxed acquisition, which does not contradict the robustness of final
regret for \gls{rr} as implemented in Ax.

The neighbour replacement speeds up only the plain loop, in which \mRRnb is 5.0 evaluations faster
than \gls{rr} ($p_\mathrm{HB}=0.019$) and as fast as enumeration (\gls{rope} mass 0.72). With Ax's
settings, it is as fast as \gls{rr} (\gls{rope} mass 0.97), whereas enumeration is faster than both
(posterior probability 0.96 against \gls{rr} and 0.95 against \mRRnb, $p_\mathrm{HB}=0.28$ for both), and all three end at practically
equivalent regret (\gls{rope} mass at least 0.98). Holm--Bonferroni correction is over the three pairs
of each setting and metric. We ran this loop only at $d=2$, so both the effect of the restart
selection and that of the replacement remain to be tested in higher dimensions.

\subsection{Per-Task Results}
\label{app:pertask}
\Cref{tab:pertaskA,tab:pertaskB} report the final normalized regret of seven methods on every task,
so that the mean ranks of \cref{tab:main} can be checked against the magnitudes. On Tier A, the \gls{gp}
methods reach zero regret on most $d=2$ tasks, and their differences come from the $d=4$ and $d=8$
tasks. On Tier B, no single task dominates the ranks.

\begin{table}[p]
\caption{Final normalized regret per task on Tier A (counted view, mean over 30 seeds and half-width
of the 95\% $t$ interval, best mean in \textbf{bold}). The noise level is the standard deviation
of the observation noise as a fraction of the range between the optimum and the reference value.}
\label{tab:pertaskA}
\centering
\scriptsize
\setlength{\tabcolsep}{3pt}
\input{tables/per_task_A.tex}
\end{table}

\begin{table}[p]
\caption{Final normalized regret per task on Tier B (counted view, mean over 20 seeds and half-width
of the 95\% $t$ interval, best mean in \textbf{bold}).}
\label{tab:pertaskB}
\centering
\scriptsize
\setlength{\tabcolsep}{3pt}
\input{tables/per_task_B.tex}
\end{table}

\subsection{JAHS-Bench-201}
\label{app:jahs}
On the three JAHS tasks (13 mixed dimensions with 8 categorical inputs, 20 seeds), \mAxDef ranks
first on every task and \mPR eleventh on average. With three tasks, the Nemenyi \gls{cd} (11.45) separates
only \mAxDef from random search and Sobol. Duplicate rates are negligible for all methods (at most 0.3\%), because
the continuous and categorical inputs make exact duplicates rare.

\subsection{CIFAR-10 Encodings and Training Precision}
\label{app:cifar}
\noindent\textbf{Encodings.}
With three runs per encoding for \mRRAx, \mDisc and \mAxDef, \mRRAx re-proposes 0--18 trainings with $K=6$, 25--32
with $K=24$, 28--34 with power-of-two widths and none with a log-continuous learning rate, while
\mDisc and \mAxDef re-propose none. The best observed validation accuracies of the three methods, averaged
over runs, end within 0.4~pp of each other on every encoding. On the coarse encodings \mRRAx is even
slightly ahead (e.g., 81.9\% vs.\ 81.6\% with power-of-two widths), since its duplicate trainings
draw new seeds and the best observed value favors lucky draws.
\begin{figure}[t]
\centering
\includegraphics{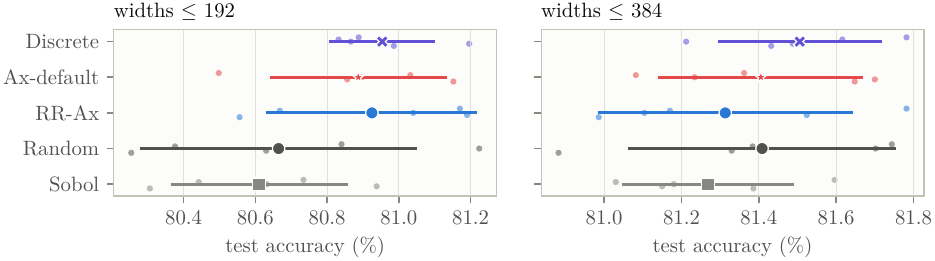}
\caption{\textbf{CIFAR-10 Final Accuracy.} Test accuracy of the final configuration of each run (small dots) and mean
$\pm$ one standard deviation over the 5 runs (large markers).}
\label{fig:cifar}
\end{figure}

\noindent\textbf{Training Precision.}
We retrained the final configuration of seed 0 of every space and method with the same five
seeds in bf16 autocast and in fp32 (50 pairs). The difference is $-0.03$~pp in validation accuracy
(95\% CI $[-0.16, 0.07]$, $p=0.85$) and $-0.01$~pp in test accuracy ($[-0.15, 0.12]$, $p=0.98$). Since fp32 is 1.6 to 2.1$\times$ slower, we measure costs in number of trainings.

\subsection{Accuracy versus Network Size}
\label{app:moo}
\begin{figure}[h]
\centering
\includegraphics[width=\textwidth]{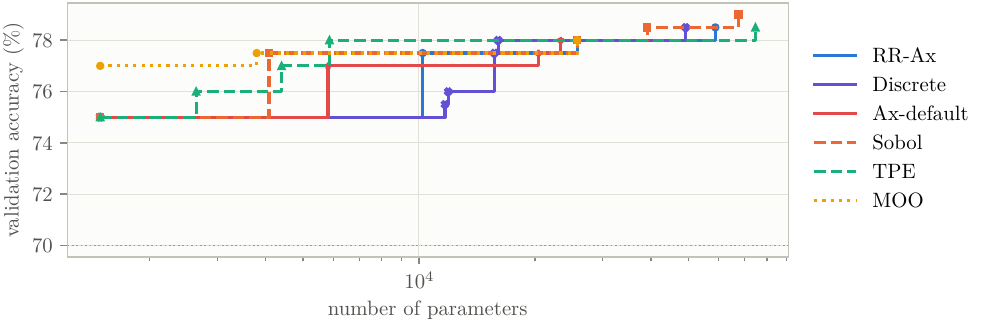}\\[2mm]
\includegraphics[width=\textwidth]{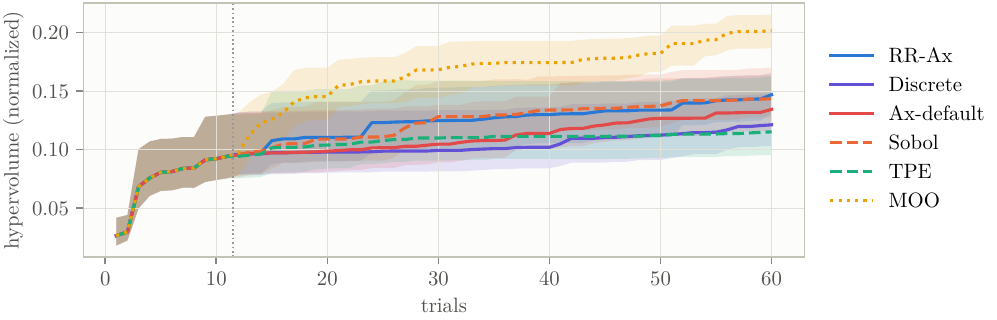}
\caption{\textbf{Accuracy versus Network Size.} Accuracy versus size on credit-g: pooled Pareto fronts over 10 seeds (\textbf{\emph{top}}) and
hypervolume as trials accumulate (\textbf{\emph{bottom}}, mean and 95\% \acrshort{bca} band).}
\label{fig:pareto}
\end{figure}

As an extension, we ran multi-objective \gls{bo} with the batch log Noisy Expected Hypervolume
Improvement (qLogNEHVI, \citealp{daulton2021parallel, ament2023unexpected}) on the validation
accuracy and parameter count of a two-layer \gls{mlp} on the
OpenML credit-g dataset (two integer widths on a $K=16$ grid, batch size as a power of two, two
log-scaled floats, 60 trials, 10 seeds), and compared the Pareto fronts of all methods in a
normalized space (accuracy from the majority-class rate to 100\%, log parameter count between the
smallest and the largest network). As expected, since the single-objective methods ignore the
size, the multi-objective run reaches the largest hypervolume, 0.202 (95\% CI $[0.187, 0.216]$),
against 0.147 for \mRRAx, 0.143 for Sobol, 0.134 for \mAxDef, 0.121 for \mDisc and
0.115 for \gls{tpe} ($p_\mathrm{HB}\le0.012$ against every method), and finds a 77\% accurate network
with 1{,}489 parameters where \mAxDef needs 5{,}809 (\cref{fig:pareto}). As on JAHS-Bench-201, no
method re-proposes a configuration here, since exact duplicates would also require the two continuous
hyperparameters to coincide.

\section{Implementation and Reproducibility}
\label{app:impl}

\subsection{Software and Baselines}
\label{app:software}
We use Python 3.12, Ax 1.3.1, BoTorch 0.18.1 and PyTorch 2.11 with CUDA 12.8, with all versions
pinned in a lock file. The Tier B libraries run in isolated environments behind a JSON-lines bridge, since
their pinned dependencies conflict with the Ax stack. Every run writes its trial log, the
resolved configuration, the code revision and the software environment.

\noindent\textbf{Reproducing \mRRAx in Current Ax.}
Ax $\geq$1.0 keeps integer parameters discrete. \mRRAx reproduces the earlier behavior through the
legacy transform list that relaxes integers (\texttt{IntToFloat}), which still ships with Ax,
while \mDisc raises the enumeration threshold of Ax from $10^4$ to $2\times10^6$ grid points.

\noindent\textbf{Grid Representation.}
Integer hyperparameters are represented by grid indices, and each benchmark maps indices to
physical values (linear step, power of two or lookup table). Duplicates are detected on the
canonical index tuple after rounding.

\noindent\textbf{Baselines.}
\Cref{tab:baselines} lists the library, configuration and integer handling of every baseline. All
of them receive the shared Sobol design as their first $n_0$ trials and are told every observation.

\begin{table}[h]
\caption{Baseline configurations. Integers enter every method as grid indices $\{0,\dots,n-1\}$.}
\label{tab:baselines}
\centering
\scriptsize
\setlength{\tabcolsep}{3pt}
\begin{tabular}{@{}p{1.3cm}p{2.3cm}p{5.6cm}p{2.3cm}p{2.6cm}@{}}
\toprule
Method & Library & Configuration & Integers & Initial design \\
\midrule
Random & own & uniform sampling with replacement & indices & first $n_0$ trials \\
Sobol & own & continues the shared scrambled Sobol sequence & indices & same sequence \\
\gls{tpe} & Optuna 5.0.0 & \texttt{TPESampler}, multivariate, $n_0$ startup trials & \texttt{IntDistribution} & \texttt{enqueue\_trial} \\
\gls{smac} & SMAC3 2.4.1 & hyperparameter optimization facade, deterministic, one call per configuration & \texttt{Integer} & replaces SMAC's initial design \\
TuRBO & own (TuRBO-1) & Thompson sampling, length $0.8$ in $[0.5^7, 1.6]$, failure tolerance $\max(4,d)$, success tolerance 3 & relax and round & told as data \\
HEBO & HEBO 0.3.6 & \gls{gp}, MACE acquisition, NSGA-II (population and iterations 100), duplicates rejected & relax and round & observed before the first proposal \\
\gls{cocabo} & own, BoTorch mixed \gls{gp} & EXP3 bandit per discrete dimension, \gls{ucb} ($\beta=2$) for the continuous ones & categorical arms & told, updating the bandits \\
\mAxDef & Ax 1.3.1 & \texttt{choose\_generation\_strategy} (``fast''), default model and acquisition, deduplication & discrete & attached, count toward the Sobol stage \\
\bottomrule
\end{tabular}
\end{table}

\subsection{\acrshort{bo} Loop Settings}
\label{app:settings}
The relaxed variants maximize the acquisition with BoTorch's multi-start L-BFGS-B (20 restarts
from the best of 1{,}024 raw samples). Ax passes its rounding to this optimizer, which rounds the end
point of every restart and keeps the restart with the highest acquisition after rounding. In fully
discrete spaces, Ax enumerates up to $10^4$ combinations, uses a discrete local search up to 100
values per dimension and otherwise BoTorch's alternating optimizer (20 restarts, 1{,}024 raw samples,
at most 64 alternations, dimensions with more than 20 values relaxed). Ax normalizes the inputs to the
unit cube, winsorizes the worst tail of the outcomes and standardizes them. The kernel has the
dimension-scaled prior $\mathrm{LogNormal}(\sqrt2+\tfrac12\log d,\sqrt3)$ on each lengthscale with a
floor of 0.025, and the inferred noise has a $\mathrm{LogNormal}(-4,1)$ prior with a floor of $10^{-4}$
(standardized units). The noise-free ablation reports a standard error of zero, which gives a
fixed-noise likelihood clamped at $10^{-6}$. Hyperparameters are refit by maximum marginal likelihood
after every observation. A batch of $q=4$ is built sequentially, each point conditioned on the pending
ones. The pure-BoTorch loop of \mPR and \mRRnb uses the same kernel, priors and likelihood with
standardized outcomes but without winsorization, and the BoTorch default sampler for \gls{qlognei}.

\noindent\textbf{Probabilistic Reparameterization (\mPR).}
\mPR is our reimplementation of the Monte Carlo estimator of \citet{daulton2022bayesian}: each integer dimension with
relaxed value $\theta$ is drawn as $\lfloor\theta\rfloor$ plus a Bernoulli variable of mean
$\theta-\lfloor\theta\rfloor$, the expectation of \acrshort{qlognei} is estimated with 64 draws, its gradient
with respect to $\theta$ with the likelihood-ratio estimator and a mean baseline, and $\theta$ is
optimized with Adam (50 steps from the best 8 of 256 random starts). The proposal is the best of
1{,}024 draws from the optimized distribution, without excluding observed points. With 128 draws
and 100 steps the optimized acquisition is the same at about three times the cost. An analytic
variant that enumerates the $2^m$ corners of the rounding cell with \acrshort{logei} ranks 4.64
and 5.48 on Tiers A and B in place of the Monte Carlo estimator (5.47 and 4.72) and re-proposes 18\% of its trials on Tier A, at 146\,s per proposal
on JAHS-Bench-201 ($2^{13}$ corners). Both results therefore apply to our implementation of \acrshort{pr} and may not hold for others.

\noindent\textbf{Fitted Lengthscales.}
After every \gls{bo} fit, we record the \gls{ard} lengthscales in the normalized model space and convert them
to grid steps: one step spans $1/(n-1)$ of the normalized range for discrete transforms and $1/n$
for the relaxed ones, whose bounds are widened to $[-\tfrac12, n-\tfrac12]$.

\subsection{Benchmarks}
\label{app:bench}
Tier A uses the BoTorch implementations of Rosenbrock, Ackley, Levy and Styblinski-Tang on their
default domains ($[-5,10]^d$, $[-32.768,32.768]^d$, $[-10,10]^d$ and $[-5,5]^d$). Each dimension has
$n=16$ levels with step $(u-l)/(n-1)$, shifted by less than half a step so that the global
optimum $x^*$ is a grid point, i.e., the levels are $x^*_j+(i-k^*_j)\,\Delta_j$ with
$k^*_j=\mathrm{round}((x^*_j-l_j)/\Delta_j)$. In the granularity study, $n=192/K$. The observation noise is
Gaussian with standard deviation $0.018$ or $0.06$ times the range between the optimum and the
reference value, i.e., about 0.3 and 1~pp on the accuracy scale of our CIFAR-10 search, and is
drawn with a seed per evaluation. LCBench runs on YAHPO Gym~1.0.2 \citep{pfisterer2022yahpo} at
the maximal fidelity of 52 epochs and maximizes validation accuracy. Its seven hyperparameters are
batch size $[16,512]$, maximal units $[64,1024]$ and number of layers $[1,5]$ (integers on unit
steps), the learning rate $[10^{-4},10^{-1}]$ (log scale), and maximal dropout, momentum and weight
decay (linear). The 15 instances are 3945, 7593, 34539, 126025, 126026, 126029, 146212, 167152,
167168, 167181, 167200, 168329, 168868, 189866 and 189905: the first five were our initial selection,
and the other ten a seeded random sample of the remaining 29. The granularity study and the batch
experiment ($q=4$, 10 seeds, budget 100) use 3945, 7593 and 34539. HPOBench (version 0.0.10)
provides the tabular \gls{mlp} benchmark on OpenML tasks 31, 53, 3917, 9952 and 10101 at its default
fidelity (243 iterations, full training set), with five ordinal hyperparameters (depth, width,
batch size, $L_2$ penalty and learning rate) encoded as grid indices, and minimizes one minus validation
accuracy. JAHS-Bench-201 (version 0.1.0, surrogate benchmark) runs on CIFAR-10, Fashion-MNIST and
Colorectal Histology at 200 epochs and maximizes validation accuracy over $N$, $W$ and the resolution
(three levels each, as grid indices), eight categorical inputs and two log-scaled floats. Tier B
uses a budget of 100 distinct configurations, $n_0=2d+1$ and 20 seeds.

\subsection{CIFAR-10 Training}
\label{app:cifartrain}
The \gls{cnn} stacks four $5\times5$ convolutions (stride 1, padding 1), each followed by batch
normalization and ReLU, with one $2\times2$ max-pooling after the second, and a single linear layer on
the flattened $10\times10$ feature map. It is trained with Adam ($\beta=(0.9,0.999)$, $L_2$ weight
decay $10^{-4}$), a constant learning rate, a deliberately small batch size of 10, cross-entropy, no data augmentation and
inputs scaled to $[-1,1]$. The 45k/5k split is stratified (500 validation images per class, fixed
seed), and the weights of the epoch with the best validation accuracy are kept. Each training draws
its own seed for initialization and batch order. The 20 configurations of the 15-epoch check are
the first 20 points of a scrambled Sobol design on the grid, each trained once for 25 epochs. Training
ran in bf16 autocast on an NVIDIA RTX~5060~Ti and in fp32 on an NVIDIA T4 (15 of the 50 main runs).
The final evaluation retrains the three best distinct configurations of each run with five new
seeds and keeps the one with the highest mean validation accuracy.

\subsection{Budget Cap and Cost}
\label{app:cost}
\noindent\textbf{Budget Cap.}
A run that does not reach $B$ distinct configurations within $2B$ trials is censored, and its
regret in the unique view is taken at the cap. On Tier A, this happens to 21\% of the runs of
\mRRAx, 20\% of \mRRcache, 13\% of \mKRound, 7\% of \mRRext and \mPR and 52\% of TuRBO; on Tier B, to 24\% of
\mRRAx and \mRRcache, 16\% of \mDisc (mixed spaces, \cref{sec:boundary}), 10\% of \mKRound, 8\% of \mRRext,
3\% of \mPR and 23\% of TuRBO. No other method reaches the cap. In the ablations, \gls{ucb} with
$\beta=0.5$ and $\beta=2$ reaches it in 99\% and 76\% of the runs of \mRRAx and $\beta=0.5$ in 98\% of
those of \mDisc without exclusion, so their unique-view regret is truncated, whereas the counted
view of \cref{tab:ablation} is not.

\noindent\textbf{Computational Cost.}
A proposal takes 0.1\,s for \mDisc and \mAxDef at $d=2$, against 0.6\,s for \mRRAx and 1.4\,s for \mKRound. At $d=4$,
enumerating the 65{,}536 grid points makes \mDisc slower (3.2\,s vs.\ 0.4\,s for \mAxDef), and at $d=8$ \mDisc
falls back to the alternating optimizer (0.9\,s vs.\ 6.5\,s for \mRRAx). \mPR needs about 22\,s per proposal
on JAHS-Bench-201 against 5--7\,s for the other variants, and HEBO, which we run on Tiers A and B
only, 2.4--8.2\,s on Tier A.

\noindent\textbf{Compute.}
Tiers A and B ran on CPUs. A CIFAR-10 \gls{bo} run (50 trainings of 15 epochs) plus its 15
re-evaluation trainings took 5 to 10 hours on one shared \gls{gpu} in bf16 with widths up to 384,
i.e., 5 to 9 minutes per training.

\end{document}

%% file: tables/main.tex
\begin{tabular}{lrrrrrr}
\toprule
 & \multicolumn{3}{c}{mean rank of final regret} & \multicolumn{3}{c}{duplicate rate (\%)} \\
\cmidrule(lr){2-4}\cmidrule(lr){5-7}
method & A (36) & B (20) & JAHS (3) & A & B & JAHS \\
\midrule
RR-Ax & 5.28 & 7.12 & 6.00 & 10.9 & 16.9 & 0.0 \\
RR-ext & 5.47 & 4.08 & 4.00 & 13.2 & 10.2 & 0.3 \\
RR+cache & 5.92 & 7.12 & 6.00 & 10.7 & 16.9 & 0.0 \\
RR+dedup & 5.79 & 6.10 & 6.00 & 0.0 & 0.0 & 0.0 \\
KernelRound & 4.17 & 6.85 & 2.33 & 5.7 & 7.3 & 0.0 \\
PR & 5.47 & 4.72 & 10.33 & 4.2 & 3.7 & 0.0 \\
Discrete & 4.39 & 3.48 & 4.67 & 0.0 & 7.6 & 0.1 \\
Ax-default & 5.03 & 3.58 & 1.00 & 0.0 & 0.0 & 0.0 \\
\midrule
HEBO & 6.04 & 4.60 & -- & 0.0 & 0.0 & -- \\
TuRBO & 9.81 & 9.95 & 9.67 & 45.3 & 12.1 & 0.0 \\
TPE & 9.50 & 10.80 & 6.00 & 3.5 & 2.2 & 0.0 \\
SMAC & 11.86 & 10.60 & 10.00 & 0.0 & 0.0 & 0.0 \\
CoCaBO & 13.81 & 12.10 & 12.33 & 4.3 & 0.1 & 0.0 \\
Random & 13.56 & 14.90 & 13.00 & 3.7 & 0.0 & 0.0 \\
Sobol & 13.92 & 14.00 & 13.67 & 0.0 & 0.0 & 0.0 \\
\midrule
Nemenyi CD & 3.57 & 4.80 & 11.45 & & & \\
\bottomrule
\end{tabular}

%% file: tables/ablation.tex
\begin{tabular}{lrrrr}
\toprule
variant & dup.\ (\%) & $\Delta$ dup.\ (pp) & $p_{\mathrm{HB}}$ & $\Delta$ regret ($p_{\mathrm{HB}}$) \\
\midrule
RR-Ax (qLogNEI, Mat\'ern 5/2, inferred noise) & 2.6 & -- & -- & -- \\
\quad noise fixed to 0 & \textbf{5.0} & \textbf{+2.4} & 0.024 & +0.047 (0.633) \\
\quad EI & \textbf{11.3} & \textbf{+8.7} & $<$0.001 & +0.010 (1.000) \\
\quad UCB $\beta=0.5$ & \textbf{62.6} & \textbf{+59.9} & $<$0.001 & \textbf{+0.068} (0.002) \\
\quad UCB $\beta=2$ & \textbf{41.4} & \textbf{+38.8} & $<$0.001 & -0.020 (1.000) \\
\quad UCB $\beta=8$ & 4.1 & +1.5 & 0.604 & +0.025 (0.059) \\
\quad RBF kernel & \textbf{2.1} & \textbf{-0.5} & 0.046 & -0.003 (1.000) \\
Discrete & 0.0 & -- & -- & -- \\
\quad noise fixed to 0 & 0.0 & +0.0 & 1.000 & +0.020 (0.515) \\
\quad EI & 0.0 & +0.0 & 1.000 & +0.005 (1.000) \\
\quad RBF kernel & 0.0 & +0.0 & 1.000 & +0.001 (1.000) \\
\quad Discrete, no excl., UCB $\beta=0.5^\dagger$ & \textbf{58.3} & \textbf{+58.3} & $<$0.001 & \textbf{+0.097} ($<$0.001) \\
\bottomrule
\end{tabular}

%% file: tables/delay.tex
\begin{tabular}{@{}l@{\hspace{4pt}}r@{\hspace{5pt}}r@{\hspace{5pt}}r@{}}
\toprule
Ax & A & B & A, $d{=}2$ \\
\midrule
\mRRAx & 61.4 & 84.7 & 17.3 \\
\mRRdedup & 62.8 & 83.9 & 18.9 \\
\mDisc & 57.4 & 79.9 & 15.2 \\
\mDisc, no excl. & -- & -- & 15.0 \\
\midrule
BoTorch, $d{=}2$ & & plain & Ax \\
\midrule
\acrshort{rr} &  & 21.5 & 19.9 \\
\mRRnb &  & 16.5 & 19.3 \\
Enumeration &  & 16.9 & 16.4 \\
\bottomrule
\end{tabular}

%% file: tables/cifar.tex
\begin{tabular}{lrrrrrrrr}
\toprule
 & \multicolumn{3}{c}{widths $\leq 192$} & \multicolumn{3}{c}{widths $\leq 384$} & \multicolumn{2}{c}{$\Delta$ vs.\ Sobol (paired)} \\
\cmidrule(lr){2-4}\cmidrule(lr){5-7}\cmidrule(lr){8-9}
method & val. & test & dup. & val. & test & dup. & val. & test \\
\midrule
Discrete & \textbf{81.89 $\pm$ 0.14} & \textbf{80.95 $\pm$ 0.15} & 0.0 & \textbf{82.41 $\pm$ 0.16} & \textbf{81.51 $\pm$ 0.21} & 0.0 & \textbf{+0.28 (10)} & \textbf{+0.29 (10)} \\
Ax-default & 81.86 $\pm$ 0.16 & 80.89 $\pm$ 0.25 & 0.0 & 82.34 $\pm$ 0.16 & 81.41 $\pm$ 0.27 & 0.0 & +0.23 (9) & +0.21 (7) \\
RR-Ax & 81.89 $\pm$ 0.19 & 80.92 $\pm$ 0.29 & 14.6 & 82.27 $\pm$ 0.25 & 81.31 $\pm$ 0.33 & 2.0 & +0.21 (7) & +0.18 (8) \\
Random & 81.64 $\pm$ 0.29 & 80.66 $\pm$ 0.39 & 0.0 & 82.27 $\pm$ 0.30 & 81.41 $\pm$ 0.35 & 0.0 & +0.09 (8) & +0.10 (6) \\
Sobol & 81.54 $\pm$ 0.27 & 80.61 $\pm$ 0.25 & 0.0 & 82.20 $\pm$ 0.20 & 81.27 $\pm$ 0.22 & 0.0 & -- & -- \\
\bottomrule
\end{tabular}

%% file: tables/rope.tex
\begin{tabular}{llccc}
\toprule
Tier & Pair (a vs.\ b) & ROPE 0.005 & ROPE 0.01 & ROPE 0.02 \\
\midrule
A & RR-Ax vs.\ Ax-default & 0.11 / 0.76 / 0.13 & 0.00 / 0.99 / 0.01 & 0.00 / 1.00 / 0.00 \\
A & RR-Ax vs.\ Discrete & 0.00 / 0.63 / 0.37 & 0.00 / 0.97 / 0.03 & 0.00 / 1.00 / 0.00 \\
A & Discrete vs.\ Ax-default & 0.22 / 0.78 / 0.00 & 0.01 / 0.99 / 0.00 & 0.00 / 1.00 / 0.00 \\
\midrule
B & RR-Ax vs.\ Ax-default & 0.00 / 0.10 / 0.90 & 0.00 / 0.45 / 0.55 & 0.00 / 0.95 / 0.05 \\
B & RR-Ax vs.\ Discrete & 0.00 / 0.11 / 0.89 & 0.00 / 0.49 / 0.51 & 0.00 / 0.99 / 0.01 \\
B & Discrete vs.\ Ax-default & 0.00 / 0.99 / 0.01 & 0.00 / 1.00 / 0.00 & 0.00 / 1.00 / 0.00 \\
\bottomrule
\end{tabular}

%% file: tables/loop.tex
\begin{tabular}{@{}lrrrr@{}}
\toprule
Loop & Dup.\ (\%) & Stalled (\%) & Regret & Evaluations \\
\midrule
\acrshort{rr}, plain & 64.7 & 57.5 & 0.035 & 21.5 \\
\quad + best restart after rounding & 21.3 & 5.8 & 0.017 & 19.1 \\
\quad + cells of width $1/n$ & 64.6 & 60.8 & 0.052 & 21.6 \\
\quad + winsorization & 70.0 & 50.0 & 0.068 & 21.8 \\
\quad + cells and winsorization & 67.9 & 45.8 & 0.067 & 22.6 \\
\quad + all three & 22.9 & 1.7 & 0.012 & 19.9 \\
\mRRAx (Ax) & 24.5 & 2.2 & 0.007 & 17.3 \\
\midrule
\mRRnb, plain & 0.0 & 0.0 & 0.015 & 16.5 \\
\mRRnb, Ax settings & 0.0 & 0.0 & 0.005 & 19.3 \\
Enumeration, plain & 0.0 & 0.0 & 0.006 & 16.9 \\
Enumeration, Ax settings & 0.0 & 0.0 & 0.006 & 16.4 \\
\bottomrule
\end{tabular}

%% file: tables/per_task_A.tex
\begin{tabular}{@{}lrrrrrrr@{}}
\toprule
task & \mRRAx & \mDisc & \mAxDef & HEBO & \acrshort{tpe} & \acrshort{smac} & Random \\
\midrule
Ackley, $d{=}2$, noise 0 & \textbf{0.000{\tiny$\pm$0.000}} & \textbf{0.000{\tiny$\pm$0.000}} & \textbf{0.000{\tiny$\pm$0.000}} & \textbf{0.000{\tiny$\pm$0.000}} & \textbf{0.000{\tiny$\pm$0.000}} & 0.262{\tiny$\pm$0.110} & 0.444{\tiny$\pm$0.097} \\
Ackley, $d{=}2$, noise 0.018 & \textbf{0.000{\tiny$\pm$0.000}} & \textbf{0.000{\tiny$\pm$0.000}} & \textbf{0.000{\tiny$\pm$0.000}} & \textbf{0.000{\tiny$\pm$0.000}} & 0.034{\tiny$\pm$0.049} & 0.253{\tiny$\pm$0.113} & 0.444{\tiny$\pm$0.097} \\
Ackley, $d{=}2$, noise 0.06 & 0.022{\tiny$\pm$0.045} & \textbf{0.000{\tiny$\pm$0.000}} & \textbf{0.000{\tiny$\pm$0.000}} & \textbf{0.000{\tiny$\pm$0.000}} & 0.017{\tiny$\pm$0.035} & 0.297{\tiny$\pm$0.108} & 0.446{\tiny$\pm$0.098} \\
Ackley, $d{=}4$, noise 0 & 0.017{\tiny$\pm$0.035} & \textbf{0.000{\tiny$\pm$0.000}} & \textbf{0.000{\tiny$\pm$0.000}} & \textbf{0.000{\tiny$\pm$0.000}} & 0.388{\tiny$\pm$0.056} & 0.601{\tiny$\pm$0.084} & 0.764{\tiny$\pm$0.032} \\
Ackley, $d{=}4$, noise 0.018 & 0.061{\tiny$\pm$0.073} & 0.032{\tiny$\pm$0.065} & 0.056{\tiny$\pm$0.081} & \textbf{0.013{\tiny$\pm$0.026}} & 0.345{\tiny$\pm$0.069} & 0.707{\tiny$\pm$0.057} & 0.764{\tiny$\pm$0.033} \\
Ackley, $d{=}4$, noise 0.06 & 0.369{\tiny$\pm$0.167} & 0.276{\tiny$\pm$0.163} & 0.280{\tiny$\pm$0.165} & \textbf{0.233{\tiny$\pm$0.133}} & 0.433{\tiny$\pm$0.076} & 0.771{\tiny$\pm$0.062} & 0.799{\tiny$\pm$0.040} \\
Ackley, $d{=}8$, noise 0 & 0.058{\tiny$\pm$0.045} & \textbf{0.000{\tiny$\pm$0.000}} & \textbf{0.000{\tiny$\pm$0.000}} & 0.091{\tiny$\pm$0.103} & 0.420{\tiny$\pm$0.054} & 0.782{\tiny$\pm$0.038} & 0.886{\tiny$\pm$0.022} \\
Ackley, $d{=}8$, noise 0.018 & 0.119{\tiny$\pm$0.095} & \textbf{0.032{\tiny$\pm$0.066}} & 0.086{\tiny$\pm$0.094} & 0.236{\tiny$\pm$0.135} & 0.528{\tiny$\pm$0.096} & 0.822{\tiny$\pm$0.036} & 0.896{\tiny$\pm$0.025} \\
Ackley, $d{=}8$, noise 0.06 & 0.675{\tiny$\pm$0.128} & \textbf{0.651{\tiny$\pm$0.165}} & 0.700{\tiny$\pm$0.145} & 0.742{\tiny$\pm$0.117} & 0.926{\tiny$\pm$0.053} & 0.942{\tiny$\pm$0.029} & 0.924{\tiny$\pm$0.029} \\
Levy, $d{=}2$, noise 0 & 0.004{\tiny$\pm$0.007} & \textbf{0.000{\tiny$\pm$0.000}} & \textbf{0.000{\tiny$\pm$0.000}} & \textbf{0.000{\tiny$\pm$0.000}} & 0.002{\tiny$\pm$0.002} & 0.017{\tiny$\pm$0.011} & 0.043{\tiny$\pm$0.016} \\
Levy, $d{=}2$, noise 0.018 & \textbf{0.001{\tiny$\pm$0.002}} & 0.005{\tiny$\pm$0.003} & 0.008{\tiny$\pm$0.003} & 0.004{\tiny$\pm$0.003} & 0.011{\tiny$\pm$0.006} & 0.027{\tiny$\pm$0.012} & 0.045{\tiny$\pm$0.017} \\
Levy, $d{=}2$, noise 0.06 & \textbf{0.009{\tiny$\pm$0.005}} & 0.031{\tiny$\pm$0.014} & 0.033{\tiny$\pm$0.015} & 0.022{\tiny$\pm$0.011} & 0.031{\tiny$\pm$0.015} & 0.044{\tiny$\pm$0.019} & 0.059{\tiny$\pm$0.020} \\
Levy, $d{=}4$, noise 0 & \textbf{0.003{\tiny$\pm$0.002}} & 0.004{\tiny$\pm$0.003} & 0.008{\tiny$\pm$0.009} & 0.014{\tiny$\pm$0.011} & 0.017{\tiny$\pm$0.010} & 0.051{\tiny$\pm$0.017} & 0.073{\tiny$\pm$0.016} \\
Levy, $d{=}4$, noise 0.018 & 0.014{\tiny$\pm$0.005} & \textbf{0.008{\tiny$\pm$0.003}} & 0.010{\tiny$\pm$0.004} & 0.031{\tiny$\pm$0.016} & 0.020{\tiny$\pm$0.005} & 0.057{\tiny$\pm$0.015} & 0.076{\tiny$\pm$0.016} \\
Levy, $d{=}4$, noise 0.06 & 0.035{\tiny$\pm$0.008} & \textbf{0.024{\tiny$\pm$0.008}} & 0.026{\tiny$\pm$0.007} & 0.046{\tiny$\pm$0.016} & 0.051{\tiny$\pm$0.011} & 0.081{\tiny$\pm$0.020} & 0.085{\tiny$\pm$0.018} \\
Levy, $d{=}8$, noise 0 & 0.007{\tiny$\pm$0.003} & 0.008{\tiny$\pm$0.005} & \textbf{0.006{\tiny$\pm$0.002}} & 0.039{\tiny$\pm$0.017} & 0.052{\tiny$\pm$0.024} & 0.084{\tiny$\pm$0.013} & 0.146{\tiny$\pm$0.017} \\
Levy, $d{=}8$, noise 0.018 & 0.018{\tiny$\pm$0.004} & 0.025{\tiny$\pm$0.008} & \textbf{0.015{\tiny$\pm$0.003}} & 0.055{\tiny$\pm$0.014} & 0.060{\tiny$\pm$0.017} & 0.092{\tiny$\pm$0.012} & 0.152{\tiny$\pm$0.020} \\
Levy, $d{=}8$, noise 0.06 & 0.039{\tiny$\pm$0.010} & 0.034{\tiny$\pm$0.007} & \textbf{0.033{\tiny$\pm$0.009}} & 0.086{\tiny$\pm$0.019} & 0.084{\tiny$\pm$0.019} & 0.134{\tiny$\pm$0.021} & 0.176{\tiny$\pm$0.023} \\
Rosenbrock, $d{=}2$, noise 0 & 0.000{\tiny$\pm$0.000} & \textbf{0.000{\tiny$\pm$0.000}} & \textbf{0.000{\tiny$\pm$0.000}} & 0.000{\tiny$\pm$0.000} & 0.000{\tiny$\pm$0.000} & 0.000{\tiny$\pm$0.000} & 0.002{\tiny$\pm$0.001} \\
Rosenbrock, $d{=}2$, noise 0.018 & 0.004{\tiny$\pm$0.002} & 0.005{\tiny$\pm$0.002} & 0.004{\tiny$\pm$0.002} & 0.005{\tiny$\pm$0.001} & \textbf{0.004{\tiny$\pm$0.001}} & 0.005{\tiny$\pm$0.002} & 0.004{\tiny$\pm$0.002} \\
Rosenbrock, $d{=}2$, noise 0.06 & 0.016{\tiny$\pm$0.007} & 0.012{\tiny$\pm$0.005} & \textbf{0.008{\tiny$\pm$0.004}} & 0.009{\tiny$\pm$0.004} & 0.012{\tiny$\pm$0.006} & 0.018{\tiny$\pm$0.010} & 0.014{\tiny$\pm$0.008} \\
Rosenbrock, $d{=}4$, noise 0 & 0.000{\tiny$\pm$0.000} & 0.000{\tiny$\pm$0.000} & 0.000{\tiny$\pm$0.000} & \textbf{0.000{\tiny$\pm$0.000}} & 0.000{\tiny$\pm$0.000} & 0.003{\tiny$\pm$0.001} & 0.007{\tiny$\pm$0.002} \\
Rosenbrock, $d{=}4$, noise 0.018 & 0.006{\tiny$\pm$0.002} & 0.005{\tiny$\pm$0.002} & \textbf{0.005{\tiny$\pm$0.002}} & 0.006{\tiny$\pm$0.002} & 0.008{\tiny$\pm$0.002} & 0.013{\tiny$\pm$0.004} & 0.011{\tiny$\pm$0.004} \\
Rosenbrock, $d{=}4$, noise 0.06 & 0.018{\tiny$\pm$0.006} & 0.016{\tiny$\pm$0.005} & \textbf{0.013{\tiny$\pm$0.005}} & 0.014{\tiny$\pm$0.004} & 0.022{\tiny$\pm$0.009} & 0.021{\tiny$\pm$0.006} & 0.032{\tiny$\pm$0.012} \\
Rosenbrock, $d{=}8$, noise 0 & 0.000{\tiny$\pm$0.000} & 0.000{\tiny$\pm$0.000} & 0.001{\tiny$\pm$0.000} & \textbf{0.000{\tiny$\pm$0.000}} & 0.001{\tiny$\pm$0.000} & 0.011{\tiny$\pm$0.006} & 0.033{\tiny$\pm$0.007} \\
Rosenbrock, $d{=}8$, noise 0.018 & 0.007{\tiny$\pm$0.002} & 0.007{\tiny$\pm$0.002} & 0.007{\tiny$\pm$0.002} & \textbf{0.005{\tiny$\pm$0.001}} & 0.008{\tiny$\pm$0.003} & 0.026{\tiny$\pm$0.009} & 0.035{\tiny$\pm$0.008} \\
Rosenbrock, $d{=}8$, noise 0.06 & 0.013{\tiny$\pm$0.003} & 0.015{\tiny$\pm$0.004} & \textbf{0.012{\tiny$\pm$0.004}} & 0.023{\tiny$\pm$0.007} & 0.026{\tiny$\pm$0.007} & 0.049{\tiny$\pm$0.011} & 0.051{\tiny$\pm$0.015} \\
Styblinski-Tang, $d{=}2$, noise 0 & 0.006{\tiny$\pm$0.013} & 0.008{\tiny$\pm$0.017} & \textbf{0.000{\tiny$\pm$0.000}} & \textbf{0.000{\tiny$\pm$0.000}} & 0.018{\tiny$\pm$0.020} & 0.072{\tiny$\pm$0.040} & 0.161{\tiny$\pm$0.036} \\
Styblinski-Tang, $d{=}2$, noise 0.018 & 0.006{\tiny$\pm$0.009} & \textbf{0.000{\tiny$\pm$0.000}} & \textbf{0.000{\tiny$\pm$0.000}} & \textbf{0.000{\tiny$\pm$0.000}} & 0.027{\tiny$\pm$0.026} & 0.050{\tiny$\pm$0.033} & 0.162{\tiny$\pm$0.037} \\
Styblinski-Tang, $d{=}2$, noise 0.06 & \textbf{0.016{\tiny$\pm$0.014}} & 0.031{\tiny$\pm$0.018} & 0.046{\tiny$\pm$0.026} & 0.041{\tiny$\pm$0.022} & 0.025{\tiny$\pm$0.024} & 0.103{\tiny$\pm$0.042} & 0.168{\tiny$\pm$0.039} \\
Styblinski-Tang, $d{=}4$, noise 0 & 0.046{\tiny$\pm$0.019} & \textbf{0.039{\tiny$\pm$0.019}} & 0.052{\tiny$\pm$0.026} & 0.059{\tiny$\pm$0.023} & 0.138{\tiny$\pm$0.034} & 0.145{\tiny$\pm$0.025} & 0.242{\tiny$\pm$0.027} \\
Styblinski-Tang, $d{=}4$, noise 0.018 & 0.053{\tiny$\pm$0.022} & 0.059{\tiny$\pm$0.019} & \textbf{0.045{\tiny$\pm$0.025}} & 0.061{\tiny$\pm$0.022} & 0.126{\tiny$\pm$0.032} & 0.154{\tiny$\pm$0.032} & 0.243{\tiny$\pm$0.027} \\
Styblinski-Tang, $d{=}4$, noise 0.06 & \textbf{0.098{\tiny$\pm$0.029}} & 0.102{\tiny$\pm$0.024} & 0.117{\tiny$\pm$0.024} & 0.106{\tiny$\pm$0.028} & 0.166{\tiny$\pm$0.035} & 0.175{\tiny$\pm$0.038} & 0.265{\tiny$\pm$0.032} \\
Styblinski-Tang, $d{=}8$, noise 0 & 0.089{\tiny$\pm$0.014} & \textbf{0.083{\tiny$\pm$0.018}} & 0.096{\tiny$\pm$0.025} & 0.098{\tiny$\pm$0.024} & 0.176{\tiny$\pm$0.023} & 0.167{\tiny$\pm$0.022} & 0.336{\tiny$\pm$0.022} \\
Styblinski-Tang, $d{=}8$, noise 0.018 & 0.103{\tiny$\pm$0.022} & \textbf{0.098{\tiny$\pm$0.017}} & 0.122{\tiny$\pm$0.019} & 0.111{\tiny$\pm$0.028} & 0.188{\tiny$\pm$0.025} & 0.188{\tiny$\pm$0.023} & 0.341{\tiny$\pm$0.023} \\
Styblinski-Tang, $d{=}8$, noise 0.06 & 0.177{\tiny$\pm$0.026} & \textbf{0.163{\tiny$\pm$0.014}} & 0.209{\tiny$\pm$0.026} & 0.169{\tiny$\pm$0.027} & 0.234{\tiny$\pm$0.025} & 0.254{\tiny$\pm$0.023} & 0.356{\tiny$\pm$0.027} \\
\bottomrule
\end{tabular}

%% file: tables/per_task_B.tex
\begin{tabular}{@{}lrrrrrrr@{}}
\toprule
task & \mRRAx & \mDisc & \mAxDef & HEBO & \acrshort{tpe} & \acrshort{smac} & Random \\
\midrule
HPOBench NN 10101 & 0.156{\tiny$\pm$0.043} & \textbf{0.029{\tiny$\pm$0.012}} & 0.031{\tiny$\pm$0.011} & 0.067{\tiny$\pm$0.029} & 0.077{\tiny$\pm$0.034} & 0.169{\tiny$\pm$0.054} & 0.293{\tiny$\pm$0.051} \\
HPOBench NN 31 & 0.081{\tiny$\pm$0.012} & \textbf{0.048{\tiny$\pm$0.012}} & 0.055{\tiny$\pm$0.011} & 0.054{\tiny$\pm$0.010} & 0.062{\tiny$\pm$0.013} & 0.081{\tiny$\pm$0.014} & 0.111{\tiny$\pm$0.015} \\
HPOBench NN 3917 & 0.038{\tiny$\pm$0.007} & 0.036{\tiny$\pm$0.005} & 0.037{\tiny$\pm$0.007} & 0.035{\tiny$\pm$0.005} & \textbf{0.033{\tiny$\pm$0.008}} & 0.044{\tiny$\pm$0.009} & 0.069{\tiny$\pm$0.010} \\
HPOBench NN 53 & 0.016{\tiny$\pm$0.004} & 0.010{\tiny$\pm$0.004} & 0.008{\tiny$\pm$0.003} & \textbf{0.007{\tiny$\pm$0.003}} & 0.011{\tiny$\pm$0.004} & 0.013{\tiny$\pm$0.003} & 0.032{\tiny$\pm$0.004} \\
HPOBench NN 9952 & 0.023{\tiny$\pm$0.007} & \textbf{0.007{\tiny$\pm$0.002}} & 0.008{\tiny$\pm$0.003} & 0.011{\tiny$\pm$0.002} & 0.015{\tiny$\pm$0.003} & 0.023{\tiny$\pm$0.006} & 0.055{\tiny$\pm$0.014} \\
LCBench 126025 & 0.092{\tiny$\pm$0.036} & 0.099{\tiny$\pm$0.037} & \textbf{0.077{\tiny$\pm$0.008}} & 0.220{\tiny$\pm$0.075} & 0.411{\tiny$\pm$0.078} & 0.212{\tiny$\pm$0.075} & 0.447{\tiny$\pm$0.046} \\
LCBench 126026 & \textbf{0.014{\tiny$\pm$0.003}} & 0.015{\tiny$\pm$0.002} & 0.014{\tiny$\pm$0.001} & 0.018{\tiny$\pm$0.028} & 0.260{\tiny$\pm$0.076} & 0.094{\tiny$\pm$0.057} & 0.332{\tiny$\pm$0.056} \\
LCBench 126029 & 0.109{\tiny$\pm$0.026} & \textbf{0.072{\tiny$\pm$0.015}} & 0.078{\tiny$\pm$0.020} & 0.136{\tiny$\pm$0.032} & 0.271{\tiny$\pm$0.056} & 0.219{\tiny$\pm$0.042} & 0.364{\tiny$\pm$0.034} \\
LCBench 146212 & 0.012{\tiny$\pm$0.002} & 0.012{\tiny$\pm$0.004} & 0.013{\tiny$\pm$0.003} & \textbf{0.008{\tiny$\pm$0.002}} & 0.021{\tiny$\pm$0.007} & 0.021{\tiny$\pm$0.005} & 0.087{\tiny$\pm$0.023} \\
LCBench 167152 & 0.005{\tiny$\pm$0.001} & \textbf{0.004{\tiny$\pm$0.001}} & 0.004{\tiny$\pm$0.001} & 0.009{\tiny$\pm$0.004} & 0.036{\tiny$\pm$0.006} & 0.033{\tiny$\pm$0.005} & 0.049{\tiny$\pm$0.004} \\
LCBench 167168 & 0.062{\tiny$\pm$0.010} & 0.053{\tiny$\pm$0.001} & 0.060{\tiny$\pm$0.012} & \textbf{0.009{\tiny$\pm$0.003}} & 0.105{\tiny$\pm$0.030} & 0.061{\tiny$\pm$0.027} & 0.216{\tiny$\pm$0.025} \\
LCBench 167181 & 0.026{\tiny$\pm$0.021} & \textbf{0.002{\tiny$\pm$0.001}} & 0.002{\tiny$\pm$0.001} & 0.035{\tiny$\pm$0.069} & 0.156{\tiny$\pm$0.088} & 0.057{\tiny$\pm$0.030} & 0.271{\tiny$\pm$0.044} \\
LCBench 167200 & 0.152{\tiny$\pm$0.023} & 0.142{\tiny$\pm$0.010} & 0.147{\tiny$\pm$0.011} & \textbf{0.078{\tiny$\pm$0.062}} & 0.468{\tiny$\pm$0.080} & 0.263{\tiny$\pm$0.092} & 0.641{\tiny$\pm$0.038} \\
LCBench 168329 & 0.177{\tiny$\pm$0.030} & 0.155{\tiny$\pm$0.025} & 0.137{\tiny$\pm$0.019} & \textbf{0.101{\tiny$\pm$0.034}} & 0.442{\tiny$\pm$0.038} & 0.210{\tiny$\pm$0.060} & 0.481{\tiny$\pm$0.043} \\
LCBench 168868 & 0.018{\tiny$\pm$0.006} & 0.006{\tiny$\pm$0.001} & 0.006{\tiny$\pm$0.001} & \textbf{0.005{\tiny$\pm$0.002}} & 0.044{\tiny$\pm$0.012} & 0.036{\tiny$\pm$0.019} & 0.098{\tiny$\pm$0.020} \\
LCBench 189866 & 0.122{\tiny$\pm$0.059} & 0.082{\tiny$\pm$0.007} & \textbf{0.076{\tiny$\pm$0.000}} & 0.079{\tiny$\pm$0.082} & 0.384{\tiny$\pm$0.082} & 0.254{\tiny$\pm$0.098} & 0.660{\tiny$\pm$0.051} \\
LCBench 189905 & 0.011{\tiny$\pm$0.003} & 0.010{\tiny$\pm$0.002} & 0.010{\tiny$\pm$0.001} & \textbf{0.008{\tiny$\pm$0.004}} & 0.050{\tiny$\pm$0.016} & 0.031{\tiny$\pm$0.007} & 0.153{\tiny$\pm$0.020} \\
LCBench 34539 & 0.001{\tiny$\pm$0.001} & \textbf{0.000{\tiny$\pm$0.000}} & 0.000{\tiny$\pm$0.000} & 0.002{\tiny$\pm$0.000} & 0.018{\tiny$\pm$0.006} & 0.006{\tiny$\pm$0.002} & 0.074{\tiny$\pm$0.014} \\
LCBench 3945 & 0.001{\tiny$\pm$0.001} & \textbf{0.000{\tiny$\pm$0.000}} & 0.001{\tiny$\pm$0.000} & 0.001{\tiny$\pm$0.000} & 0.020{\tiny$\pm$0.013} & 0.006{\tiny$\pm$0.003} & 0.058{\tiny$\pm$0.016} \\
LCBench 7593 & \textbf{0.003{\tiny$\pm$0.004}} & 0.007{\tiny$\pm$0.005} & 0.010{\tiny$\pm$0.006} & 0.027{\tiny$\pm$0.013} & 0.166{\tiny$\pm$0.072} & 0.105{\tiny$\pm$0.026} & 0.390{\tiny$\pm$0.041} \\
\bottomrule
\end{tabular}